\documentclass{article}

\ifdefined\pdfsuppressptexinfo
\fi

\usepackage{iclr2027_conference,times}

\usepackage{amsmath,amsfonts,bm}

\def\eqref#1{equation~\ref{#1}}

\def\1{\bm{1}}

\DeclareMathAlphabet{\mathsfit}{\encodingdefault}{\sfdefault}{m}{sl}
\SetMathAlphabet{\mathsfit}{bold}{\encodingdefault}{\sfdefault}{bx}{n}

\usepackage{amsmath,amssymb}
\usepackage{booktabs}
\usepackage{xcolor,colortbl}
\usepackage{array}
\usepackage{graphicx}
\usepackage{float}
\usepackage{wrapfig}
\usepackage{tikz}
\usetikzlibrary{arrows.meta}
\usepackage{url}
\usepackage{xspace}

\usepackage{xcolor}
\definecolor{lightblue}{RGB}{83,136,233}
\definecolor{darkblue}{RGB}{32,49,154}
\colorlet{modelcolor}{darkblue}
\usepackage[colorlinks, linkcolor=darkblue, urlcolor=darkblue,citecolor=darkblue]{hyperref}

\definecolor{ResultNavy}{HTML}{233B53}
\colorlet{ResultTableHeader}{ResultNavy!80}
\definecolor{ResultTeal}{HTML}{087F8C}
\definecolor{ResultHeader}{HTML}{EAF0F6}
\definecolor{ResultStripe}{HTML}{F5F7FA}
\definecolor{ResultFocus}{HTML}{E8F5F3}
\definecolor{ResultFocusStrong}{HTML}{D1EBE7}

\newcommand{\model}{\textsc{FocusVTC}\xspace}

\title{\model: Efficient and High-Performance\\
Visual Text Compression with Adaptive\\
Resolution}

\newcommand{\paperwebicon}{%
  \raisebox{-0.15em}{%
    \begin{tikzpicture}[x=1em,y=1em,line width=0.35pt]
      \draw (0,0) circle[radius=0.43];
      \draw (0,0) ellipse[x radius=0.21,y radius=0.43];
      \draw (-0.43,0) -- (0.43,0);
      \draw (-0.375,0.21) -- (0.375,0.21);
      \draw (-0.375,-0.21) -- (0.375,-0.21);
    \end{tikzpicture}}}
\newcommand{\paperresourcelink}[3]{%
  \href{#1}{\raisebox{-0.15em}{\includegraphics[height=1em]{#2}}\hspace{0.35em}#3}}

\author{%
\begin{minipage}[t]{\dimexpr\textwidth-2\tabcolsep\relax}
\centering\normalfont
\textbf{FangZhi Zhong}$^{*1,2,4}$, \textbf{Xuerui Qiu}$^{*1,2,4}$,
\textbf{Yuqi Pan}$^{1,2}$, \textbf{Ya Liu}$^{3}$, \textbf{Shaowei Gu}$^{1,2}$ \\
\textbf{Bo XU}$^{\dagger1,2}$, \textbf{Guoqi Li}$^{\dagger1,2}$\\
{\normalfont $^{1}$Institute of Automation, Chinese Academy of Sciences} \\
{\normalfont $^{2}$School of Artificial Intelligence, University of Chinese Academy of Sciences} \\
{\normalfont $^{3}$Shanghai Jiao Tong University \quad $^{4}$Zhongguancun Academy} \\[2pt]
{\small\sffamily\bfseries
\hypersetup{urlcolor=black}%
\href{https://fangzhi-zhong.github.io}{\paperwebicon\hspace{0.35em}Website}\qquad
\paperresourcelink{https://github.com/fangzhi-zhong/FoucsVTC}{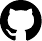}{Code}\qquad
\paperresourcelink{https://huggingface.co/zfz04/FocusVTC}{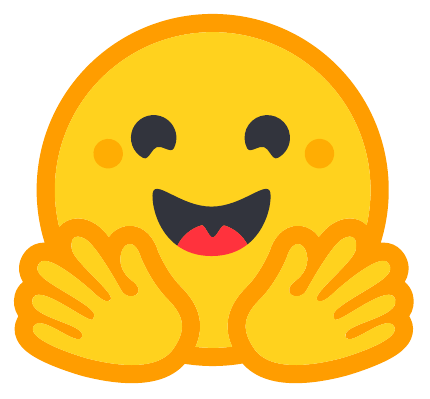}{Models}\qquad
\paperresourcelink{https://huggingface.co/datasets/zfz04/REL-CoT}{figures/link_icons/huggingface-mark.pdf}{Data}}
\end{minipage}
}

\begin{document}

\iclrfinalcopy
\maketitle
\begingroup
  \renewcommand{\thefootnote}{\fnsymbol{footnote}}
  \footnotetext[1]{Equal contribution.\quad $^{\dagger}$\,Corresponding authors.}
\endgroup
\lhead{}
\vspace{-30pt}

\begin{figure}[H]
  \centering
  \includegraphics[width=0.82\linewidth]{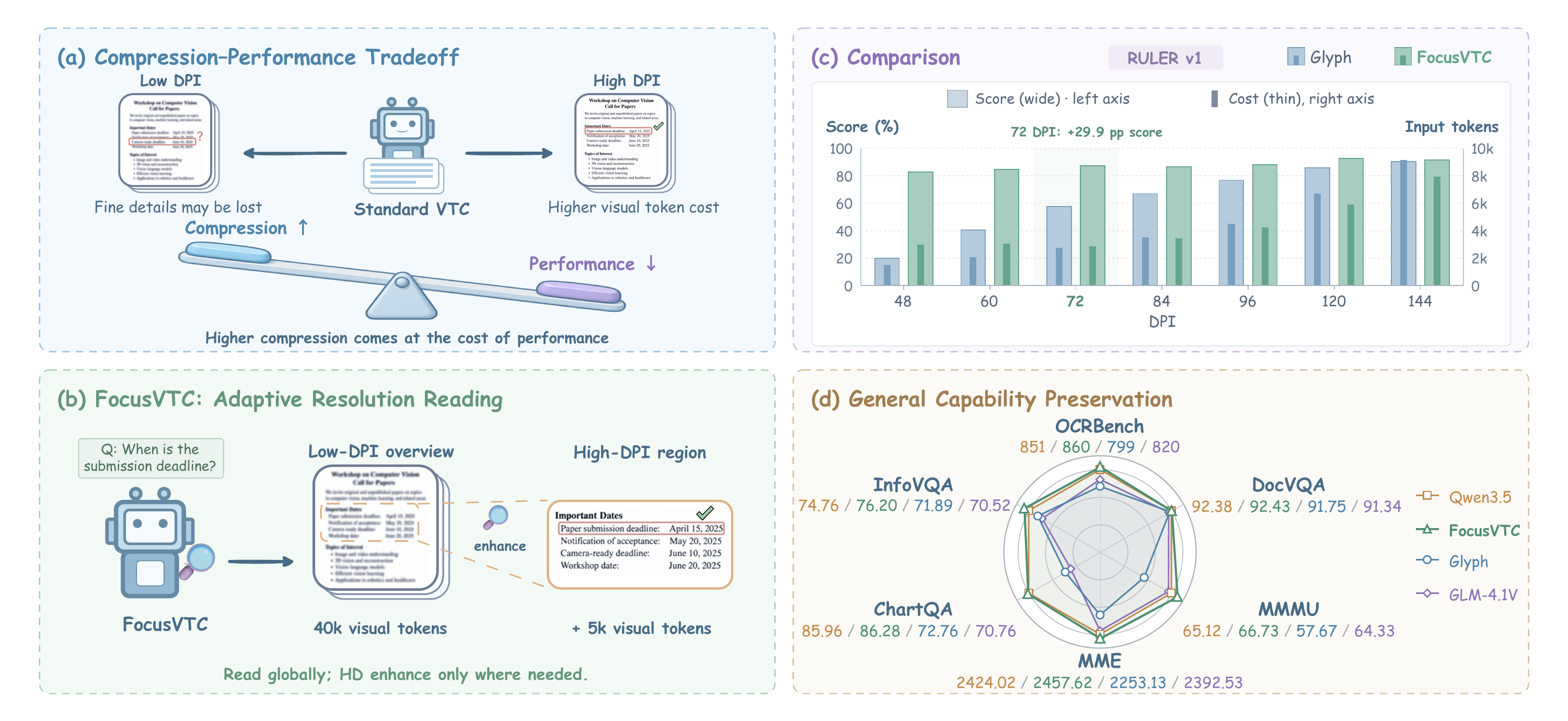}
  \caption{\textbf{Breaking the compression--performance trade-off through
  selective enhancement.} (a) Fixed-resolution VTC ties local legibility to
  whole-page token cost. (b) \model retains a compressed global view and
  selectively enhances relevant regions. (c) On RULER v1, it sustains high
  scores at low initial DPI, while Glyph needs higher DPI and more tokens
  for comparable performance. (d) These gains coexist with preserved general
  capabilities across six benchmarks.}
  \label{fig:introduction}
\end{figure}

\begin{abstract}
Long-context reasoning in large language models incurs substantial
computation and memory costs.
Visual text compression (VTC) reduces input length by rendering text as images,
but fixed-resolution rendering creates a compression--performance trade-off:
low DPI saves tokens at the expense of legibility, whereas high DPI spends
tokens on irrelevant content. We introduce \model, which breaks this trade-off
through adaptive resolution while preserving general multimodal capabilities.
It combines compressed low-DPI global views with selective region enhancement,
integrating enhanced views into ongoing reasoning.
We construct 29.4K high-quality Reasoning--Evidence Localization (REL)
chain-of-thought examples (REL-CoT) that link reasoning traces to page indices
and bounding boxes. Multi-resolution REL supervised fine-tuning (REL-SFT)
teaches the model to localize relevant regions, and Group Relative Policy
Optimization learns when to enhance resolution and how to use the resulting
observations, without a separate continual-pretraining stage.
At 72 DPI on RULER v1, \model scores 87.4 at $2.9\times$ input compression, including tool observations,
versus 57.5 for Glyph at $3.0\times$ input compression.
It surpasses its text-input backbone on
LongBench (56.40 versus 55.86), improves the MRCR macro-average by 13.91 points,
and achieves a 51.19 macro-average on VTCBench.
The MRCR latency evaluation also shows a $2.79\times$ online end-to-end
speedup over Text.
General multimodal capabilities are preserved, with MMMU increasing from
65.12 to 66.73 and MME from 2424.02 to 2457.62.

\end{abstract}


\section{Introduction}
\label{sec:introduction}
\vspace{-2mm}
Large language models (LLMs) increasingly support document analysis,
multi-document question answering, and extended interaction histories, all of
which require reasoning over long contexts \citep{yang2025qwen25million,grattafiori2024llama3,qwen2026qwen35}. As context length grows,
attention computation, inference latency, and key-value (KV) cache memory
become substantial costs \citep{dao2022flashattention,kwon2023pagedattention}; a larger context window also does not guarantee
reliable retrieval and reasoning~\citep{hsieh2024ruler,vodrahalli2024michelangelo}.
Existing approaches address different aspects of these challenges through
context-window extension~\citep{peng2024yarn},
efficient attention~\citep{dao2022flashattention},
KV cache management~\citep{kwon2023pagedattention},
and prompt compression~\citep{jiang2023llmlingua}.
Complementing these approaches, visual text compression (VTC) represents
text as page images that vision-language models (VLMs) encode with compact
visual-token sequences, preserving document coverage with fewer input tokens.

However, VTC introduces a compression--performance trade-off. Prior work~\citep{wang2024visincontext,xing2025vist} encodes long text as compact visual tokens. Glyph~\citep{cheng2025glyph} optimizes dense text
rendering, while DeepSeek-OCR~\citep{wei2025deepseekocr} explores optical
context compression. With fixed-resolution
visual input, local readability remains tied to the global token budget:
low DPI saves tokens but can obscure exact characters, whereas high DPI
improves readability by spending tokens across the entire page, including
irrelevant passages. Our key insight is that global context coverage and
precise local reading need not use the same resolution: a question often
depends on only a small subset of the document.  

We introduce \model, an
adaptive-resolution framework that breaks this trade-off by retaining a
compressed low-DPI global view and selectively enhancing relevant regions
during reasoning (Figure~\ref{fig:introduction}). \model learns this adaptive reading strategy through two-stage training. We construct 29.4K high-quality
Reasoning--Evidence Localization (REL) chain-of-thought examples (REL-CoT),
each linking a reasoning trace and answer to page indices and bounding boxes.
Multi-resolution REL supervised fine-tuning (REL-SFT) teaches the model to
localize relevant regions across seven rendering resolutions. Group Relative Policy Optimization (GRPO)~\citep{shao2024deepseekmath}
learns when and where to enhance resolution, how to integrate enhanced
views into ongoing reasoning, and when to stop. 

Experiments show strong performance with compressed visual input while preserving general multimodal capabilities. At 72 DPI on RULER v1 \citep{hsieh2024ruler}, \model scores 87.4 at $2.9\times$ input compression, including tool observations, versus 57.5 for Glyph~\citep{cheng2025glyph} at $3.0\times$ input compression. On LongBench \citep{bai2024longbench}, it outperforms its text-input backbone (56.40 vs. 55.86). On MRCR, the macro-average over six length bins and two/four/eight needles improves from 31.65 to 45.56 (+13.91).
The MRCR latency evaluation shows a $2.79\times$ online end-to-end speedup
(Appendix~\ref{app:online-latency}).
On VTCBench~\citep{zhao2025vtcbench}, \model achieves a 51.19 macro-average across Retrieval, Reasoning, and Memory. General multimodal performance is preserved, with MMMU \citep{yue2023mmmu} improving from 65.12 to 66.73 and MME \citep{fu2023mme} from 2424.02 to 2457.62. At 72 DPI, DejaVu Sans improves LongBench from 54.93 to 56.40 over Verdana (+1.47).
 Our contributions are threefold:
\begingroup
\setlength{\topsep}{0pt}
\setlength{\partopsep}{0pt}
\setlength{\itemsep}{\parskip}
\setlength{\parsep}{0pt}
\settowidth{\labelwidth}{\labelitemi}
\setlength{\leftmargini}{\labelwidth}
\addtolength{\leftmargini}{\labelsep}
\begin{itemize}
  \item We introduce an adaptive-resolution framework that breaks the compression–performance trade-off in VTC by combining compressed global context with tool-mediated access to high-resolution content for selected document regions.
  \item We construct REL-CoT, a dataset of 29.4K high-quality examples linking
        reasoning traces to page indices and bounding boxes. These annotations
        support multi-resolution REL-SFT and the learning of adaptive visual
        reading.
  \item We develop \model, which scores 87.4 on RULER v1 at
        $2.9\times$ compression and 56.40 on LongBench, improves the MRCR macro-average by 13.91
        points, and achieves 51.19 on VTCBench, while preserving general
        multimodal capabilities.
\end{itemize}
\endgroup

\section{Related Work}
\label{sec:related-work}

\paragraph{Long-context efficiency and visual text compression.}
Prior work explores context-window extension~\citep{peng2024yarn},
efficient attention~\citep{dao2022flashattention,yuan2025native},
cache management~\citep{kwon2023pagedattention},
retrieval~\citep{lewis2020retrieval},
summarization~\citep{xu2024recomp},
and prompt compression~\citep{jiang2023llmlingua}.
VisInContext and VIST compress textual context into visual
tokens~\citep{wang2024visincontext,xing2025vist}.
Glyph combines dense rendering with continual pre-training and
post-training~\citep{cheng2025glyph}, while DeepSeek-OCR studies
text reconstruction from compressed visual representations~\citep{wei2025deepseekocr}.
With fixed-resolution input, local legibility remains tied to the global
visual-token budget.
\model addresses this trade-off through compressed global context
and tool-mediated access to selected high-resolution regions.

\paragraph{Adaptive resolution and grounded visual reasoning.}
\citet{tang2026measuretransport} use transport cost to route between textual
and visual inputs and re-encode selected regions at higher resolution.
AGAR enlarges attention-selected text spans before a second inference pass
without model training~\citep{zeng2026agar}; SEER learns to select rendered
pages and retrieve their source text~\citep{xu2026seer}.
DeepEyes learns active visual inspection through reinforcement
learning~\citep{zheng2025deepeyes}. Complementary training approaches compress
reasoning into visual memory (VTC-R1), align visual- and text-input behavior
(SPIRAL), or transfer text-history policies to visual-history agents (CAPS)
\citep{wang2026vtcr1,liang2026spiral,fan2026caps}.
DocVAL distills validated spatial reasoning traces for document
grounding~\citep{guhaneogi2026docval}. Our 29.4K REL-CoT examples link
reasoning to page indices and bounding boxes across resolutions.
Multi-resolution REL-SFT and GRPO teach \model when and where to enhance
regions and how to integrate the resulting views into ongoing reasoning,
while retaining compressed global context and general multimodal
capabilities without continual pre-training.

\paragraph{Rendering fidelity and downstream evaluation.}
VTCBench evaluates retrieval, reasoning, and memory under visual
compression~\citep{zhao2025vtcbench}; Fico examines recognition and
understanding as visual fidelity and information density
vary~\citep{tu2026fico}. These studies motivate assessing task performance
alongside token savings. We evaluate prompt and observation token costs
and connect Qwen3.5's rendering-font
preferences to downstream scores: DejaVu Sans improves LongBench
performance over Verdana under matched settings
(Table~\ref{tab:training-stages}).

\begin{figure}[!t]
  \centering
  \includegraphics[width=\textwidth,trim=0 130bp 0 64bp,clip]{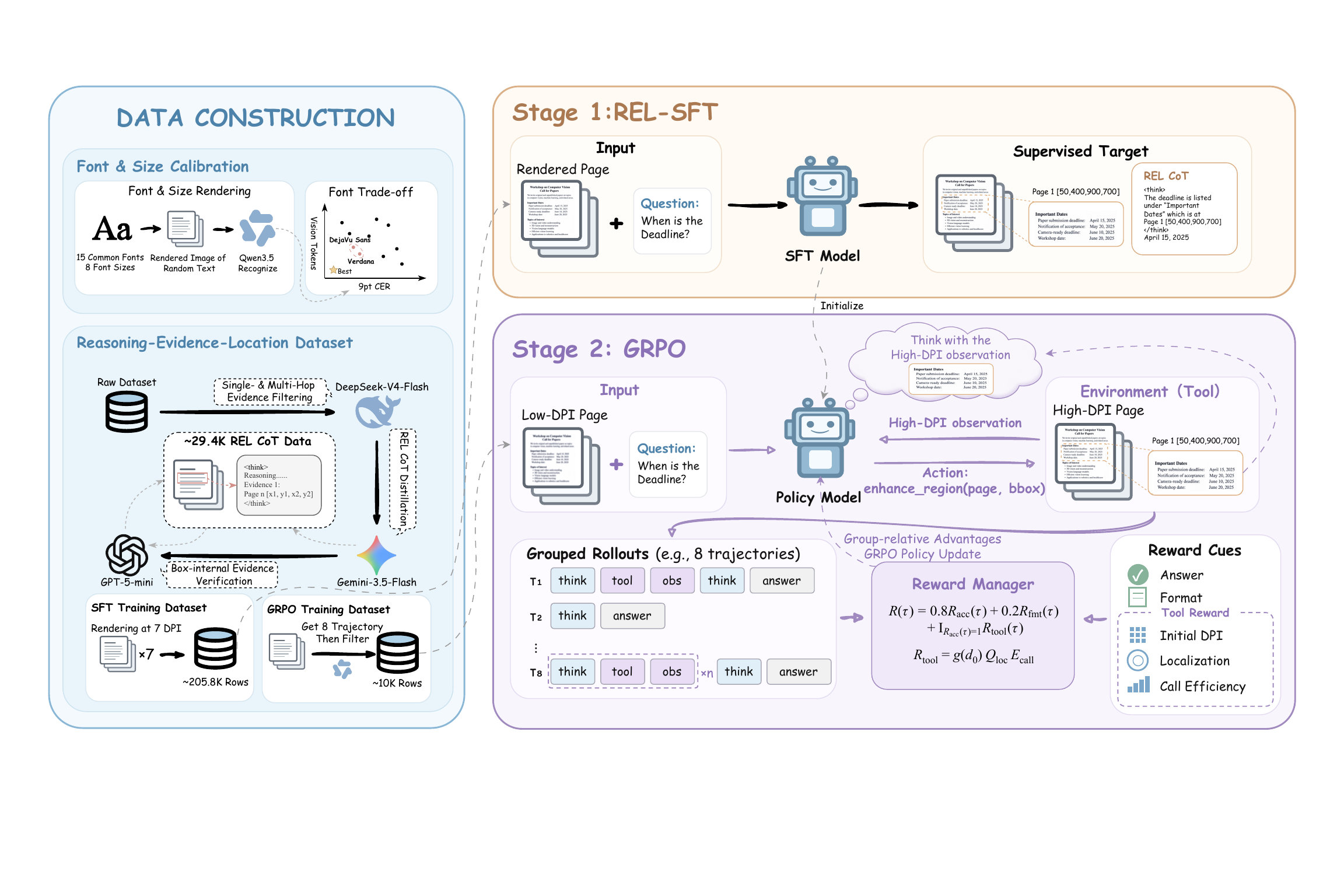}
  \caption{Overview of FocusVTC data construction and two-stage training.
  After calibrating font and point size, text contexts are rendered at
  multiple DPIs and annotated with REL-CoT. REL-SFT teaches low-DPI evidence
  localization, while GRPO trains the model to iteratively acquire and use
  high-resolution observations through selective region enhancement.}
  \label{fig:focusvtc-overview}
\end{figure}

\section{FocusVTC}
\label{sec:method}


FocusVTC learns adaptive-resolution visual text compression in two stages
(Figure~\ref{fig:focusvtc-overview}). We construct REL-CoT to provide training data for both stages, pairing each
reasoning step with a page index and a normalized evidence box.
Reasoning--Evidence Localization (REL) supervised fine-tuning uses these
annotations to teach evidence localization without continual pretraining, and
Group Relative Policy Optimization (GRPO) learns when and where to enhance
region resolution. At inference, the model reasons over low-DPI pages and
iteratively integrates enhanced views of relevant regions to produce
the final answer.

\subsection{Problem Setup}
\label{sec:problem-setup}

Let $x$ denote a source text context, and $q$ and $y$ its associated question
and answer. A renderer controlled primarily by font $f$, point size $s$, and
DPI $d$ converts $x$ into $P$ page images:
\begin{equation}
  \mathcal I^{(d)}
  =\operatorname{Render}(x;f,s,d)
  =\{I_p^{(d)}\}_{p=1}^{P}.
\end{equation}
At an initial low DPI $d_0$, the model receives the question and the global
page views. Let $r_{<t}$, $a_{<t}$, and $o_{<t}$ denote its previous reasoning
segments, actions, and tool observations, respectively (all empty at $t=0$).
The state is
\begin{equation}
  s_t=\left(q,\mathcal I^{(d_0)},r_{<t},a_{<t},o_{<t}\right),
  \qquad s_0=\left(q,\mathcal I^{(d_0)}\right).
\end{equation}
Conditioned on $s_t$, the model generates a reasoning segment $r_t$ and then
selects an action $a_t$, either a final answer or a region-enhancement call:
\begin{equation}
  (r_t,a_t)\sim\pi_\theta(\cdot\mid s_t),\qquad
  a_t\in\{\operatorname{Answer}(\hat y),
  \operatorname{Enhance\_Region}(p_t,b_t)\}.
\end{equation}
Here $p_t\in\{1,\ldots,P\}$ and $b_t\in[0,1000]^4$ specify a page and a
normalized bounding box. The tool enhances the selected region by reading
its pixels from the aligned page rendered at a higher DPI $d_h>d_0$:
\begin{equation}
  o_t=\operatorname{Enhance\_Region}\!\left(I_{p_t}^{(d_h)},b_t\right).
\end{equation}
Coordinates refer to the original page, keeping the low- and high-DPI views
aligned. After enhancement, the model appends $(r_t,a_t,o_t)$ to its history
and continues reasoning until it answers or reaches the call budget.
Figure~\ref{fig:case-study} in Appendix~\ref{app:case-study} illustrates this
iterative use of high-resolution views. The policy must localize relevant regions in the compressed global
context and enhance them selectively, balancing answer quality against the
additional visual-token and interaction cost.

\subsection{Data Construction}
\label{sec:data-rendering}

\begin{figure}[!t]
  \centering
  \includegraphics[width=\linewidth]{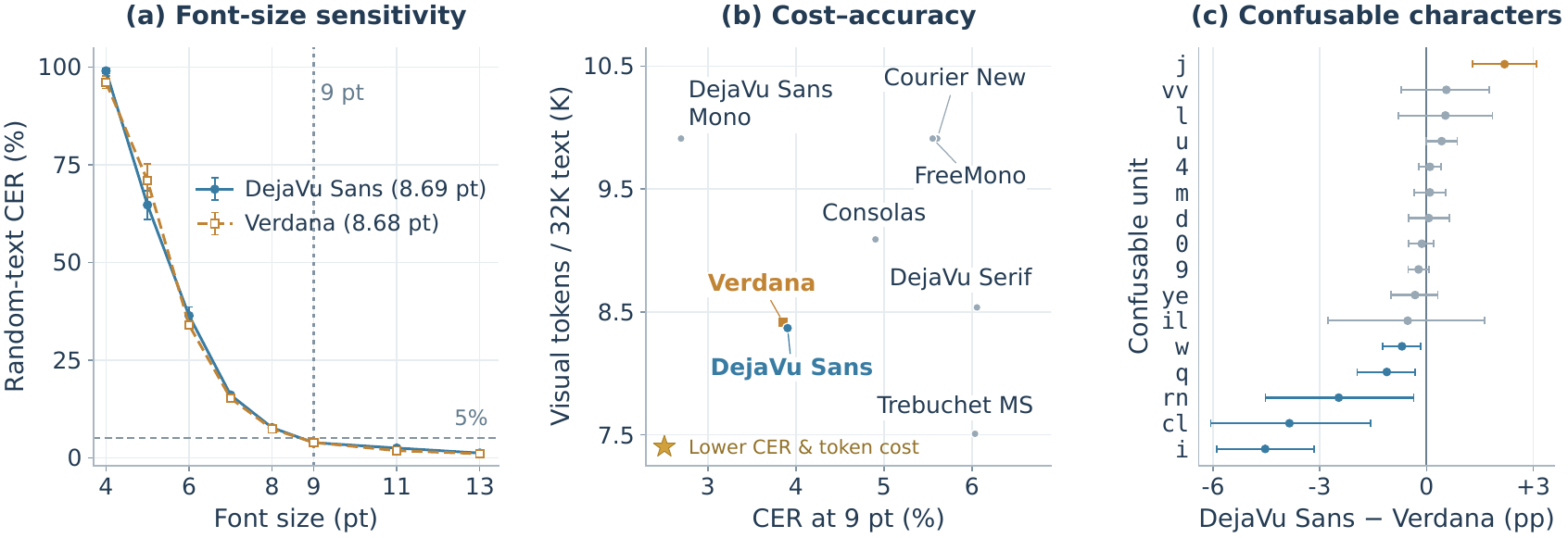}
  \caption{Font and point-size selection at 72 DPI. (a) Random-text CER
  across point sizes, with one-standard-error bars; references mark 5\% CER
  and 9~pt. (b) 9~pt cost--CER comparison; the star indicates the preferred
  direction. (c) Confusable-unit errors, DejaVu Sans minus Verdana, with
  95\% paired-bootstrap intervals.}
  \label{fig:font-selection}
\end{figure}

\noindent\textbf{Font and Point-Size Selection.}
Qwen3.5-9B exhibits rendering-font preferences that matter for compressed
reading (Figure~\ref{fig:font-selection}). We evaluate 15 fonts at eight
point sizes using randomized-text character error rate (CER), visual-token
cost, and confusable sequences. DejaVu Sans and Verdana first satisfy the
5\% CER criterion at 9~pt, with interpolated thresholds of 8.69 and
8.68~pt (panel~a). Among fonts meeting this criterion, DejaVu Sans has the
lowest cost: 8,368.4 visual tokens per 32K-token context versus 8,417.8
for Verdana (panel~b). Their overall random-text CERs are close, but DejaVu
Sans reduces CER by 0.30 percentage points on confusable sequences
(95\% CI $[-0.52,-0.09]$), with fewer errors on \texttt{cl}, \texttt{rn},
and \texttt{i}; Verdana performs better on \texttt{j} (panel~c).
We therefore use DejaVu Sans at 9~pt with 1~pt additional line spacing.
This choice also improves downstream LongBench scores from 54.93 to 56.40
(Table~\ref{tab:training-stages}). Full protocols and controls are in
Appendix~\ref{app:rendering-diagnostics}.

\noindent\textbf{Reasoning--Evidence Localization Chain-of-Thought (REL-CoT) Data Construction.}
The raw candidate pool draws primarily from the ChatQA training data
\citep{nvidia2024chatqadata} and the ChatQA2 training data
\citep{nvidia2024chatqa2data}, supplemented by TriviaQA
\citep{joshi2017triviaqa} and multi-hop and numerical-reasoning QA datasets including HotpotQA,
2WikiMultihopQA, MuSiQue, and FinQA
\citep{yang2018hotpotqa,ho2020twowiki,trivedi2022musique,chen2021finqa}.
These sources cover
reading comprehension, long contexts, tables, numerical reasoning, and
multi-hop QA. Data expansion and training details are given in
Appendix~\ref{app:training-details}.

As shown in Figure~\ref{fig:focusvtc-overview}, we use a three-stage,
cross-model pipeline so that REL-CoT examples are context-dependent, spatially
grounded, and independently verified. Before rendering, DeepSeek V4 Flash
filters out questions answerable from common knowledge, preventing shortcuts
that bypass document reading. We then render the retained contexts into pages.
Gemini 3.5 Flash serves as the annotation teacher, producing reasoning traces
with inline page indices and normalized bounding boxes. The text explains
how each selected region supports the reasoning, and a trace may refer to
multiple regions. Appendix~\ref{app:rel-case-study} shows a complete annotated
example. Finally, GPT-5 mini independently checks whether the selected
high-resolution regions support the reference answer and removes samples
with incorrect or insufficient support. The resulting release contains
approximately 29.4K verified REL-CoT examples. Appendix~\ref{app:training-data}
reports their source distribution (Figure~\ref{fig:rel-source-inventory})
and the expanded sample count.

\subsection{Reasoning--Evidence Localization Supervised Fine-Tuning}
\label{sec:sft-rel}

Reasoning--Evidence Localization supervised fine-tuning (REL-SFT) trains
FocusVTC to ground its reasoning in question-relevant document regions across
input resolutions.

For each verified REL-CoT example, we render the document at
$d_0\in\{48,60,72,84,96,120,144\}$ DPI. The seven views share the same
question $q$, annotated reasoning trace, and answer, differing only in the
input rendering (Appendix~\ref{app:training-data}). We serialize the reasoning
trace, including its page and region markers, followed by the answer into a
single target token sequence $\mathbf u$.

Given rendered pages $\mathcal I$ and question $q$, REL-SFT minimizes the
token-level negative log-likelihood
\begin{equation}
  \mathcal L
  =-\mathbb E_{(\mathcal I,q,\mathbf u)\sim\mathcal D_{\mathrm{REL}}}
  \left[\sum_{k=1}^{|\mathbf u|}
  \log p_\theta\!\left(
  u_k\mid\mathcal I,q,\mathbf u_{<k}\right)\right].
\end{equation}
This joint supervision of reasoning, evidence localization, and answer
generation establishes the foundation for the model's subsequent
adaptive-resolution capabilities. Training configuration and sequence packing
are detailed in Appendix~\ref{app:training-hyperparameters}.

\subsection{Group Relative Policy Optimization for Adaptive Resolution}
\label{sec:grpo}

Starting from the REL-SFT checkpoint, GRPO
optimizes complete interaction trajectories using task-level rewards. The
policy learns when and where to enhance resolution, how to integrate the
resulting observations, and when to terminate. Enhanced regions come from
aligned high-DPI pages while the low-DPI global context remains available.

For each initial state, we sample $G=8$ trajectories and standardize their
rewards within the group to obtain advantages $\widehat A_i$. GRPO maximizes
\begin{equation}
  \mathcal J_{\mathrm{GRPO}}(\theta)=
  \mathbb E_{i,k}\!\left[
    \min\!\left(\rho_{i,k}\widehat A_i,
    \operatorname{clip}(\rho_{i,k},1-\epsilon,1+\epsilon)
    \widehat A_i\right)\right],
\end{equation}
where $\rho_{i,k}$ is the new-to-rollout policy probability ratio for
generated token $k$ in trajectory $i$, and $\epsilon$ is the clipping
threshold. Tool observations condition the policy but are excluded from
the loss.

\noindent\textbf{GRPO Data Filtering.}
We construct the final GRPO prompts only from REL-CoT's 72-, 96-, and
144-DPI views. For each candidate, eight trajectories are generated for
construction-time filtering. We discard candidates whose eight trajectories
all repeatedly call the tool without recovering an answer, or whose eight
trajectories never call the tool. We then retain 10,000 prompts in a fixed
7:2:1 DPI ratio. The online rollout group size, prompt limits, and update
schedule are detailed in Appendix~\ref{app:rollout-protocol}.


\noindent\textbf{Reward.}
For a trajectory $\tau$, the total reward combines answer correctness, output
validity, and a correctness-gated tool-use bonus:
\begin{equation}
  R(\tau)=0.8R_{\mathrm{acc}}(\tau)+0.2R_{\mathrm{fmt}}(\tau)
  +\mathbb{I}_{R_{\mathrm{acc}}(\tau)=1}\,R_{\mathrm{tool}}(\tau).
\end{equation}
Here $R_{\mathrm{acc}}$ measures answer matching and $R_{\mathrm{fmt}}$
enforces valid outputs and tool calls. The tool bonus is enabled only for
fully correct answers. It combines resolution necessity $g(d_0)$,
localization quality $Q_{\mathrm{loc}}$, and call efficiency $E_{\mathrm{call}}$.
For $N>0$ attempted calls and $E>0$ distinct annotated regions,
\begin{equation}
  R_{\mathrm{tool}}=g(d_0)\,Q_{\mathrm{loc}}\,E_{\mathrm{call}},\quad
  g(d_0)=\operatorname{clip}\!\left(\frac{144-d_0}{144-72},0,1\right)^{\gamma},\quad
  E_{\mathrm{call}}=\min(1,E/N).
\end{equation}
The resolution weight $g$ falls from one at 72 DPI to zero at 144 DPI,
favoring enhancement when the initial view is compressed. Call efficiency
$E_{\mathrm{call}}$ stays at one for $N\leq E$ and decays as $E/N$
thereafter, discouraging redundant enhancements. Localization quality is
computed as
\begin{equation}
  Q_{\mathrm{loc}}=\frac{\sum_{(i,j)\in\mathcal M}
    \operatorname{IoU}(c_i,e_j)\min\{1,2(1-s_i)\}}{\min(N,E)},
\end{equation}
where $\mathcal M$ is a maximum-weight one-to-one matching between valid
call regions $c_i$ and boxes $e_j$ on the same page.
The requested page-area fraction $s_i$ penalizes overly broad regions,
and counting invalid attempts in $N$ penalizes malformed calls.
Both $Q_{\mathrm{loc}}$ and $E_{\mathrm{call}}$ are zero when $N=0$ or $E=0$.
Table~\ref{tab:reward-boundaries} summarizes reward behavior for representative cases.


\section{Experiments}
\label{sec:experiments}


\subsection{Experimental Setup}
\label{sec:benchmarks}
\label{sec:baselines}

We compare FocusVTC with three groups of baselines: (1) text-input models,
with Qwen3.5-9B~\citep{qwen2026qwen35} as the backbone reference;
(2) fixed-resolution multimodal models, including Qwen3-VL-8B
~\citep{bai2025qwen3vl}, Qwen3.5-9B, GLM-4.1V-9B
~\citep{vteam2025glm45vglm41vthinkingversatilemultimodal}, and
Glyph~\citep{cheng2025glyph}; and (3) training-stage variants, including
FocusVTC w/o GRPO with and without tools, a no-tools ablation of FocusVTC,
intermediate FocusVTC checkpoints, direct GRPO (FocusVTC w/o SFT), and a
tools-only Qwen3.5-9B (denoted Qwen3.5-9B+tools). We evaluate long-context performance on
LongBench~\citep{bai2024longbench}, MRCR~\citep{vodrahalli2024michelangelo,openai2025mrcr},
and VTCBench~\citep{zhao2025vtcbench}, and resolution robustness and
compression efficiency on RULER~\citep{hsieh2024ruler}.
We assess capability preservation with general multimodal benchmarks
\citep{yue2023mmmu,liu2023ocrbench,mathew2021docvqa,mathew2022infographicvqa,fu2023mme,masry2022chartqa};
complete scores are in Appendix~\ref{app:general-capabilities}.
At evaluation, Text and Vision denote text input and 72-DPI pages,
respectively. For models with tool access, the region-enhancement tool
provides aligned 144-DPI regions. All rendered pages use DejaVu Sans at
9~pt unless otherwise stated. Sampling settings and tool-use budgets are
detailed in Appendix~\ref{app:eval-protocol}.

\subsection{Long-Context Performance}
\label{sec:main-results}

\paragraph{LongBench: matching text performance with compressed input.}
FocusVTC achieves the best overall score in Table~\ref{tab:longbench13}
(56.40), slightly exceeding Qwen3.5-9B Text (55.86) and outperforming the
strongest VTC baseline, Glyph~\citep{cheng2025glyph} (52.34). More
importantly, it gains 20.54 points over the same backbone at a fixed 72 DPI.
The gains concentrate on QA and synthetic retrieval, where a small number of
passages determine the answer. This pattern supports the intended mechanism:
low-resolution pages preserve global search coverage, while learned crops
restore the exact text only where needed. Summarization changes little because
its evidence is distributed rather than localized. Few-shot results are mixed,
as demonstrations can span multiple regions.

\begin{table}[!htb]
\caption{LongBench task scores (\%), grouped by task family. Avg is the unweighted arithmetic mean of the 11 displayed non-code task scores. Vision denotes 72-DPI pages.}
\label{tab:longbench13}
\centering
\begingroup
\fontsize{8}{9.6}\selectfont
\setlength{\tabcolsep}{1.2pt}
\renewcommand{\arraystretch}{1.18}
\arrayrulecolor{ResultNavy}
\resizebox{\linewidth}{!}{%
\begin{tabular}{lc*{12}{c}}
\rowcolor{ResultTableHeader}
\textcolor{white}{\textbf{Model / stage}} & \textcolor{white}{\textbf{Input}} & \multicolumn{3}{c}{\textcolor{white}{\textbf{Single-doc QA}}} & \multicolumn{2}{c}{\textcolor{white}{\textbf{Multi-doc QA}}} & \multicolumn{2}{c}{\textcolor{white}{\textbf{Summarization}}} & \multicolumn{2}{c}{\textcolor{white}{\textbf{Few-shot}}} & \multicolumn{2}{c}{\textcolor{white}{\textbf{Synthetic}}} & \textcolor{white}{\textbf{Overall}} \\
\rowcolor{ResultHeader}
& & QP & MF-En & MF-Zh & DuR & 2Wiki & MNews & QMSum & SAMSum & Trivia & PR-En & PR-Zh & \textbf{Avg} \\
LLaMA-3.1-8B~\citep{grattafiori2024llama3} & Text & 44.56 & 44.61 & 41.26 & 19.06 & 46.67 & \textbf{25.30} & \textbf{23.28} & 35.46 & 89.12 & \underline{99.50} & 62.20 & \cellcolor{ResultHeader}48.27 \\
\rowcolor{ResultStripe}
Qwen2.5-7B~\citep{yang2025qwen25million} & Text & \underline{45.29} & 43.44 & 42.12 & 16.55 & 40.51 & \underline{24.94} & \underline{22.95} & 34.59 & 86.93 & \textbf{100.00} & \underline{98.50} & \cellcolor{ResultHeader}50.53 \\
Qwen3-8B~\citep{yang2025qwen3} & Text & 44.67 & \underline{47.73} & \underline{45.21} & \underline{21.19} & \textbf{73.92} & 21.30 & 19.60 & 35.01 & 87.98 & 97.26 & \textbf{100.00} & \cellcolor{ResultHeader}\underline{53.99} \\
\rowcolor{ResultStripe}
GLM-4-9B~\citep{glm2024chatglm} & Text & 43.75 & 45.21 & 41.23 & 20.79 & 50.89 & 24.82 & 22.84 & \underline{35.84} & \underline{90.07} & \underline{99.50} & \textbf{100.00} & \cellcolor{ResultHeader}52.27 \\
Qwen3.5-9B~\citep{qwen2026qwen35} & Text & \textbf{47.23} & \textbf{49.32} & \textbf{56.58} & \textbf{25.25} & \underline{61.82} & 23.79 & 22.28 & \textbf{38.02} & \textbf{90.16} & \textbf{100.00} & \textbf{100.00} & \cellcolor{ResultHeader}\textbf{55.86} \\
\specialrule{0.45pt}{5pt}{5pt}
\rowcolor{ResultStripe}
Qwen3-VL-8B~\citep{bai2025qwen3vl} & Vision & 27.85 & 32.17 & 26.50 & 18.00 & 37.49 & 19.47 & 20.72 & 34.10 & 87.97 & 52.50 & 6.50 & \cellcolor{ResultHeader}33.02 \\
Qwen3.5-9B~\citep{qwen2026qwen35} & Vision & 36.20 & 41.10 & 28.68 & 19.38 & 38.12 & 21.79 & \textbf{22.32} & 32.70 & \textbf{91.37} & 29.25 & 33.50 & \cellcolor{ResultHeader}35.86 \\
\rowcolor{ResultStripe}
GLM-4.1V-9B~\citep{vteam2025glm45vglm41vthinkingversatilemultimodal} & Vision & 30.85 & 33.95 & 26.92 & 15.86 & 38.33 & \underline{22.55} & 16.52 & \textbf{36.52} & 88.25 & 72.50 & 40.68 & \cellcolor{ResultHeader}38.45 \\
Glyph~\citep{cheng2025glyph} & Vision & \underline{43.77} & \underline{46.49} & \underline{36.74} & \underline{25.17} & \textbf{72.49} & 22.22 & 20.71 & \underline{34.41} & 86.61 & \textbf{100.00} & \underline{87.11} & \cellcolor{ResultHeader}\underline{52.34} \\
\rowcolor{ResultFocusStrong}
\textcolor{ResultTeal}{\textbf{FocusVTC}} (Ours) & Vision & \textbf{47.71} & \textbf{48.95} & \textbf{57.78} & \textbf{30.21} & \underline{69.04} & \textbf{23.57} & \underline{21.71} & 33.87 & \underline{91.01} & \underline{98.50} & \textbf{98.00} & \cellcolor{ResultTeal!24}\textbf{56.40} \\
\bottomrule
\end{tabular}%
}
\endgroup
\end{table}

\begin{figure}[!t]
  \centering
  \includegraphics[width=\linewidth]{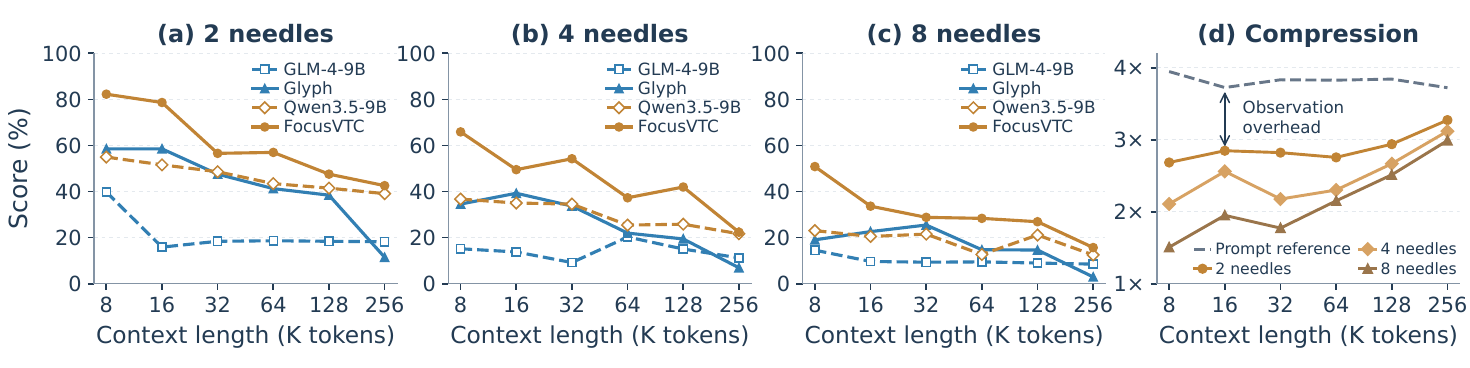}
  \caption{MRCR results: (a) two needles; (b) four needles; (c) eight needles;
  (d) prompt-plus-observation compression.
  Text is dashed/hollow; VTC is solid/filled. Context axes use bin upper
  bounds on a base-two scale. Compression divides Text tokens by the initial
  prompt plus tool observations.
  Full scores: Appendix~\ref{app:long-context-results}.}
  \label{fig:mrcr-context}
\end{figure}

\paragraph{MRCR: higher scores than text under compression.}
FocusVTC exceeds the Qwen3.5-9B Text average in all three needle settings
and also consistently outperforms Glyph, with two/four/eight-needle averages
of 60.76/45.21/30.71 versus 42.63/25.97/16.57, giving equal weight to the
six context-length bins. These gains coexist with
approximately $3.0$--$3.3\times$ prompt-plus-observation compression in the longest bin
(Figure~\ref{fig:mrcr-context}). Section~\ref{sec:efficiency} analyzes how
compression scales with context length.

\begin{table}[!t]
\centering
\caption{VTCBench performance (\%) across Retrieval, Reasoning, and Memory. Length labels are bin upper bounds (K tokens). Each Avg is the unweighted arithmetic mean of the four displayed length-bin scores. Vision denotes 72-DPI pages.}
\label{tab:vtcbench}
\label{tab:vtcbench-retrieval}
\label{tab:vtcbench-reasoning}
\label{tab:vtcbench-memory}
\begingroup
\fontsize{8}{9.6}\selectfont
\setlength{\tabcolsep}{1.2pt}
\renewcommand{\arraystretch}{1.18}
\arrayrulecolor{ResultNavy}
\resizebox{\linewidth}{!}{%
\begin{tabular}{lcccccc@{\hspace{6pt}}ccccc@{\hspace{6pt}}ccccc}
\rowcolor{ResultTableHeader}
\textcolor{white}{\textbf{Model / stage}} & \textcolor{white}{\textbf{Input}} & \multicolumn{5}{c}{\textcolor{white}{\textbf{Retrieval}}} & \multicolumn{5}{c}{\textcolor{white}{\textbf{Reasoning}}} & \multicolumn{5}{c}{\textcolor{white}{\textbf{Memory}}} \\
\rowcolor{ResultHeader}
 & & 8 & 16 & 32 & 64 & \textbf{Avg} & 8 & 16 & 32 & 64 & \textbf{Avg} & 8 & 16 & 32 & 64 & \textbf{Avg} \\
Qwen3-VL-8B~\citep{bai2025qwen3vl} & Vision & 90.50 & 77.90 & 77.01 & 76.36 & \cellcolor{ResultHeader}80.44 & 18.83 & 12.16 & 11.11 & 1.52 & \cellcolor{ResultHeader}10.91 & 21.08 & 20.18 & \underline{24.03} & \underline{20.00} & \cellcolor{ResultHeader}\underline{21.32} \\
\rowcolor{ResultStripe}
Qwen3.5-9B~\citep{qwen2026qwen35} & Vision & 89.14 & 78.45 & 79.21 & \underline{78.23} & \cellcolor{ResultHeader}81.26 & \textbf{62.97} & \textbf{55.41} & \underline{34.26} & \underline{13.64} & \cellcolor{ResultHeader}\textbf{41.57} & \underline{32.50} & \textbf{22.75} & 18.42 & 10.10 & \cellcolor{ResultHeader}20.94 \\
GLM-4.1V-9B~\citep{vteam2025glm45vglm41vthinkingversatilemultimodal} & Vision & 86.20 & 85.25 & \underline{82.76} & 58.01 & \cellcolor{ResultHeader}78.06 & 31.59 & 10.14 & 22.22 & 9.09 & \cellcolor{ResultHeader}18.26 & 25.10 & 21.89 & 13.16 & 7.07 & \cellcolor{ResultHeader}16.81 \\
\rowcolor{ResultStripe}
Glyph~\citep{cheng2025glyph} & Vision & \underline{92.08} & \underline{87.85} & 80.51 & 77.16 & \cellcolor{ResultHeader}\underline{84.40} & 22.18 & 6.08 & 15.74 & 10.61 & \cellcolor{ResultHeader}13.65 & 22.11 & 17.11 & 21.47 & 19.00 & \cellcolor{ResultHeader}19.92 \\
\rowcolor{ResultFocusStrong}
\textcolor{ResultTeal}{\textbf{FocusVTC}} (Ours) & Vision & \textbf{98.39} & \textbf{93.37} & \textbf{85.34} & \textbf{86.89} & \cellcolor{ResultTeal!24}\textbf{91.00} & \underline{45.24} & \underline{43.24} & \textbf{35.19} & \textbf{21.33} & \cellcolor{ResultTeal!24}\underline{36.25} & \textbf{32.98} & \underline{22.07} & \textbf{24.64} & \textbf{25.64} & \cellcolor{ResultTeal!24}\textbf{26.33} \\
\bottomrule
\end{tabular}%
}
\endgroup
\end{table}

\paragraph{VTCBench: retrieval, memory, and long-context reasoning.}
\label{sec:vtcbench-results}
FocusVTC obtains the best Retrieval and Memory averages in
Table~\ref{tab:vtcbench}. Its advantage is clearest in the longest bins, where
the low-resolution overview narrows the search and crops recover facts that
fixed-resolution input can miss. This supports that adaptive resolution
addresses the access-to-evidence bottleneck. Reasoning improves over the
fixed-resolution backbone in the two longest bins, although its overall
average remains lower. Overall, these results show that adaptive crops add a
complementary capability: the model can revisit and recover evidence missed
by the fixed-resolution view, with the clearest benefits at longer contexts.

\subsection{Training Strategy Analysis}
\label{sec:agent-results}

\paragraph{Training and rendering ablations.}
Table~\ref{tab:training-stages} summarizes the ablations. Training configurations
are given in Appendix~\ref{app:training-details}, and detailed results in
Appendix~\ref{app:evaluation-details}.
REL-SFT provides a useful initialization for GRPO: compared with
FocusVTC w/o SFT, FocusVTC improves eight of nine reported aggregates,
including LongBench (49.30 to 56.40), with VTCBench Reasoning as the sole
exception (38.15 to 36.25).
However, Qwen3.5-9B+tools and FocusVTC w/o GRPO+tools score lower on every
aggregate than Qwen3.5-9B and FocusVTC w/o GRPO, respectively; for
FocusVTC w/o GRPO, enabling tools reduces LongBench from 37.86 to 25.41.
After GRPO, FocusVTC surpasses FocusVTC w/o GRPO+tools on all nine
aggregates, including a rise from 22.98 to 87.38 on RULER v1, supporting
the role of policy learning in using the crop interface effectively.
FocusVTC also outperforms FocusVTC w/o tools on LongBench
(56.40 versus 36.82).
Rendering quality provides a further gain: FocusVTC with DejaVu Sans
outperforms FocusVTC (Verdana) on both RULER versions and LongBench,
with the largest RULER v1 gains
on \texttt{single\_3} and \texttt{multikey\_3}, where exact key--value
retrieval is sensitive to character errors
(Appendix~\ref{app:rendering-controls}).

\begin{table}[!t]
\caption{Training stages and tool access at 72 DPI (scores in \%).  Best and second-best scores are bold and underlined, respectively. +tools uses tool access without additional training. Detailed results are provided in Appendices~\ref{app:long-context-results} and~\ref{app:ruler-details}.}
\label{tab:training-stages}
\centering
\begingroup
\fontsize{8}{9.6}\selectfont
\setlength{\tabcolsep}{1.2pt}
\renewcommand{\arraystretch}{1.18}
\arrayrulecolor{ResultNavy}
\resizebox{\linewidth}{!}{%
\begin{tabular}{l*{12}{c}}
\rowcolor{ResultTableHeader}
\textcolor{white}{\textbf{Model}} & \textcolor{white}{\textbf{REL-SFT}} & \textcolor{white}{\textbf{GRPO}} & \textcolor{white}{\textbf{Tools}} & \multicolumn{2}{c}{\textcolor{white}{\textbf{RULER}}} & \textcolor{white}{\textbf{LongBench}} & \multicolumn{3}{c}{\textcolor{white}{\textbf{MRCR (needles)}}} & \multicolumn{3}{c}{\textcolor{white}{\textbf{VTCBench}}} \\
\rowcolor{ResultHeader}
 &  &  &  & v1 & v2 & Avg & 2 & 4 & 8 & Retrieval & Reasoning & Memory \\
\rowcolor{ResultFocusStrong}
\textcolor{ResultTeal}{\textbf{FocusVTC}} (Ours) & $\checkmark$ & $\checkmark$ & $\checkmark$ & \textbf{87.38} & \textbf{75.94} & \textbf{56.40} & \textbf{60.76} & \textbf{45.21} & \textbf{30.71} & \textbf{91.00} & 36.25 & \textbf{26.33} \\
\rowcolor{ResultFocus}
\textcolor{ResultTeal}{FocusVTC (Verdana)} (Ours) & $\checkmark$ & $\checkmark$ & $\checkmark$ & \underline{85.13} & \underline{75.49} & \underline{54.93} & \underline{60.26} & \underline{43.11} & \underline{27.34} & \underline{89.21} & 35.93 & \underline{24.23} \\
\rowcolor{ResultStripe}
FocusVTC w/o tools (Ours) & $\checkmark$ & $\checkmark$ & -- & 36.60 & 44.82 & 36.82 & 52.00 & 37.65 & 23.50 & 67.25 & 3.11 & 22.95 \\
FocusVTC w/o SFT (Ours) & -- & $\checkmark$ & $\checkmark$ & 73.21 & 62.77 & 49.30 & 57.05 & 39.36 & 26.66 & 87.17 & \underline{38.15} & 23.56 \\
\rowcolor{ResultStripe}
FocusVTC w/o GRPO (Ours) & $\checkmark$ & -- & -- & 32.28 & 43.58 & 37.86 & 44.94 & 28.83 & 21.30 & 80.27 & 20.37 & 17.68 \\
FocusVTC w/o GRPO+tools (Ours) & $\checkmark$ & -- & $\checkmark$ & 22.98 & 39.41 & 25.41 & 22.25 & 15.09 & 8.96 & 54.25 & 2.53 & 6.93 \\
\rowcolor{ResultStripe}
Qwen3.5-9B~\citep{qwen2026qwen35} & -- & -- & -- & 37.40 & 44.64 & 35.86 & 48.52 & 28.53 & 18.56 & 81.26 & \textbf{41.57} & 20.94 \\
Qwen3.5-9B+tools~\citep{qwen2026qwen35} & -- & -- & $\checkmark$ & 22.50 & 37.68 & 27.22 & 20.70 & 14.42 & 6.96 & 49.38 & 2.83 & 3.30 \\
\bottomrule
\end{tabular}}
\endgroup
\end{table}

\begin{figure}[!t]
  \centering
  \makebox[\linewidth][r]{%
    \includegraphics[width=1.03\linewidth]{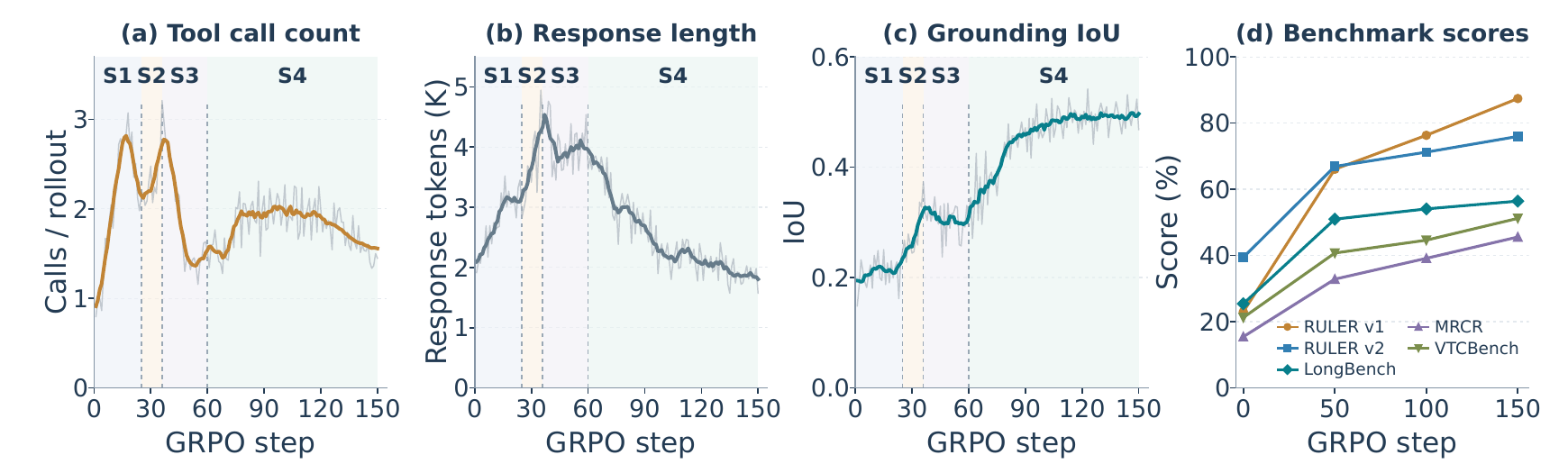}}
  \caption{Training behavior and checkpoint scores. (a) Tool calls;
  (b) response length; (c) grounding IoU; (d) benchmark scores at
  0/50/100/150 steps. 
  S1--S4 mark exploration, frequent tool use,
  fewer calls with longer responses, and shorter responses with higher IoU.}
  \label{fig:training-dynamics}
\end{figure}

\paragraph{GRPO progression.}
With tools enabled, every reported LongBench, MRCR, and
VTCBench aggregate improves steadily from 50 to 100 to 150 steps, with
particularly strong later gains on eight-needle MRCR and VTCBench Reasoning
and Memory. Figure~\ref{fig:training-dynamics} connects these gains to the
policy trajectory. In \emph{S1--S2 (0--36)},
increasing calls and response length coincide with rising IoU as the policy
learns to request useful evidence. In \emph{S3 (36--60)}, calls fall while
responses remain long and IoU briefly dips, so fewer calls alone do not imply
precise reading. In \emph{S4 (60--150)}, responses shorten and IoU rises before
stabilizing as the benchmark scores reach their
strongest checkpoint. GRPO therefore progressively converts exploratory
tool use into more selective
evidence acquisition and stopping.

\subsection{Resolution Robustness and Compression Efficiency}
\label{sec:efficiency}

\begin{figure}[!t]
  \centering
  \includegraphics[width=\linewidth]{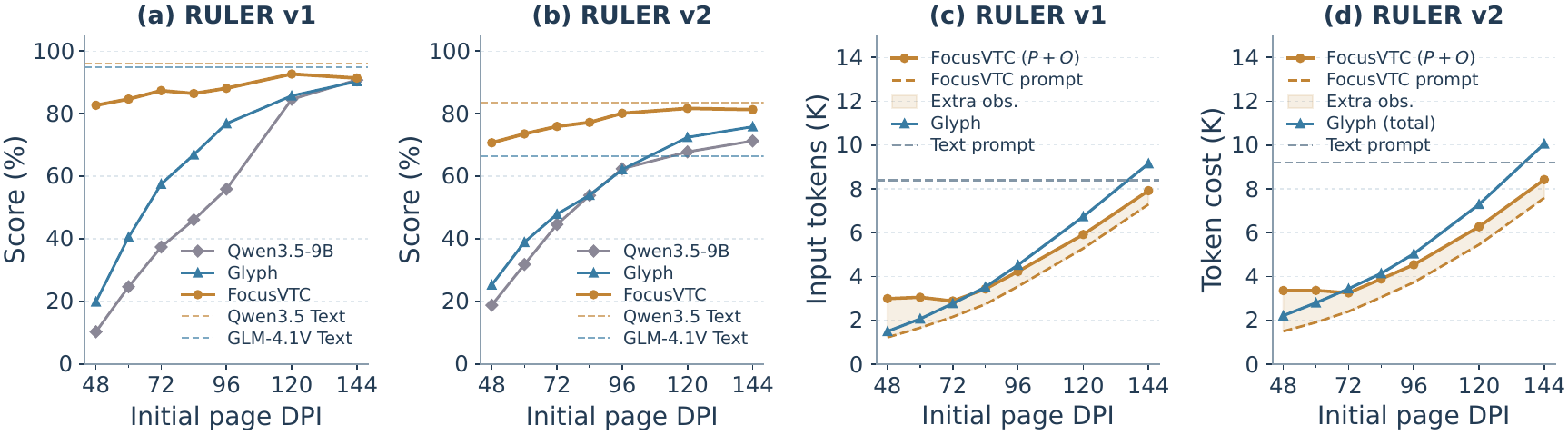}
  \caption{RULER results across DPI: (a) v1 scores; (b) v2 scores;
  (c) v1 input token costs; (d) v2 token costs.
  FocusVTC shows prompt and prompt-plus-observation costs.
  Glyph uses prompt tokens for v1 and v2. Dashed references show Text prompts.
  Token accounting: Appendix~\ref{app:cost-accounting}; task-level scores
  at 72 DPI: Appendix~\ref{app:ruler-details}.}
  \label{fig:ruler-dpi}
\end{figure}

\paragraph{Robustness across initial resolutions.}
RULER provides a direct test of the adaptive-resolution mechanism. Across
48--144 DPI, FocusVTC is substantially less sensitive to the initial
resolution than the fixed-resolution baselines (Figure~\ref{fig:ruler-dpi});
at 72 DPI, it reaches 87.38/75.94 on v1/v2, compared with 37.40/44.64 for the
same backbone without crops (Appendix~\ref{app:ruler-details}).
REL trains the model to localize across
resolutions, and the aligned 144-DPI crop then decouples exact character
reading from the initial page resolution. At 48 and 60 DPI,
extra-observation costs are higher, reducing the savings from shorter
prompts. At 72 DPI, the prompt-plus-observation length is the
smallest on both v1 and v2, while scores exceed
those at lower DPIs. Above 72 DPI, observation costs change little
but prompt length grows; starting at 144 DPI costs 2.8/2.6$\times$ as many
input tokens for only modest gains. Thus, uniformly increasing resolution spends capacity
on irrelevant text, whereas 72 DPI gives the crop policy enough information
to allocate detail selectively. We therefore use it on the other benchmarks.

\paragraph{Token costs and compression.}
At 72 DPI on RULER v1, FocusVTC uses an average of
2,154 prompt tokens and 723 extra-observation tokens. Their
sum of approximately 2,877 tokens gives $2.9\times$ compression
relative to the 8,400-token text context, while scoring 87.38.
Glyph~\citep{cheng2025glyph} scores 57.53 at $3.0\times$ compression.
FocusVTC achieves higher accuracy at
a similar compression ratio. For MRCR's longest bin, the text reference
is 197,909 tokens and the prompt is 53,181 tokens. Adding the
two/four/eight-needle observations of 7,296/10,290/13,048 tokens gives
$3.3\times/3.1\times/3.0\times$ compression
(Figure~\ref{fig:mrcr-context}(d)).
On 64K--128K four-needle MRCR examples, FocusVTC achieves a
$2.79\times$ online end-to-end speedup over Qwen3.5-9B Text, excluding
offline page rendering (Appendix~\ref{app:online-latency}).

\section{Conclusion}
\label{sec:conclusion}

Long-context reasoning incurs substantial computation and memory costs,
while fixed-resolution visual text compression trades legibility for token
savings. We presented \model, which breaks this trade-off through compressed
global views and selective high-resolution enhancement. Trained with
multi-resolution REL-SFT and GRPO on 29.4K high-quality REL-CoT examples,
\model learns to localize and enhance relevant regions without continual
pretraining. It scores 87.4 on RULER v1 at $2.9\times$
input compression, including tool observations, and surpasses its text-input
backbone on LongBench (56.40 versus 55.86). It also gains 13.91
points on MRCR and reaches a 51.19 VTCBench macro-average while preserving
general multimodal capabilities. These results show that allocating visual
detail according to reasoning needs can reconcile high compression with
strong performance. We believe this principle advances visual text
compression toward adaptive long-context understanding and provides a
foundation for more capable, versatile, and scalable multimodal intelligence.

\clearpage
\section*{AI Use Statement}
We used generative AI tools to assist with language polishing and manuscript
organization. As part of the research pipeline, we also used DeepSeek V4 Flash
to filter context-dependent questions, Gemini 3.5 Flash to generate reasoning
traces and evidence locations, and GPT-5 mini to verify whether the extracted
crops support the reference answers, as described in
Section~\ref{sec:data-rendering}.
The authors take responsibility for the final manuscript, including all
claims, experimental results, and artifacts.

\section*{Ethics Statement}
This work studies efficient long-context reading through visual text
compression and adaptive evidence localization. We use existing public QA
datasets and evaluation benchmarks to construct rendered pages and derived
reasoning--evidence annotations. The study does not recruit human participants
or collect private documents. Derived data remain subject to the usage and
redistribution conditions of their source datasets.

\section*{Reproducibility Statement}
Section~\ref{sec:method} describes the FocusVTC reading protocol, REL-CoT data
construction, REL supervised fine-tuning, and GRPO objective.
Appendix~\ref{app:training-details} reports data composition and expansion,
training hyperparameters, sequence packing, rollout limits, and representative
reward cases.
Appendix~\ref{app:evaluation-details} specifies evaluation settings and
token-cost accounting for initial prompts and tool observations, as well as
generated-token counts in the latency analysis.
Appendix~\ref{app:rendering-diagnostics} details the font
and point-size selection experiments, rendering parameters, and visual
preprocessing.

\bibliography{iclr2027_conference}
\bibliographystyle{iclr2027_conference}

\appendix
\clearpage
\section*{Appendix}
\section{Rendering Parameters and Selection}
\label{app:rendering-diagnostics}

We detail the rendering settings and font-selection diagnostics supporting
Section~\ref{sec:data-rendering} and Figure~\ref{fig:font-selection}.

\subsection{Rendering Configuration}
\label{app:visual-preprocessing}

Table~\ref{tab:rendering-parameters} lists the settings using Glyph's
parameter categories~\citep{cheng2025glyph}. Visual-token costs are measured
after preprocessing.

\begin{table}[!htb]
  \centering
  \caption{Rendering settings. The first block lists the selected font and
  resolution settings; page-layout and image-processor values in the second
  block apply to the font-selection diagnostics.}
  \label{tab:rendering-parameters}
  \small
  \begin{tabular}{@{}p{0.31\linewidth}p{0.64\linewidth}@{}}
    \toprule
    Parameter & Setting \\
    \midrule
    Font family & DejaVu Sans (selected from 15 candidates) \\
    Font size & 9\,pt (selected from 4, 5, 6, 7, 8, 9, 11, 13\,pt) \\
    Initial training DPI & 48, 60, 72, 84, 96, 120, 144 for REL-SFT;
      72, 96, 144 for GRPO \\
    Default evaluation DPI & 72, except for the resolution sweep \\
    Enhancement source DPI & 144, with aligned page layout and coordinates \\
    \midrule
    \multicolumn{2}{@{}l}{\textit{Font-selection page layout and preprocessing}} \\
    Page size & A4; $595\times842$ pixels at 72 DPI \\
    Margins & 10\,pt on each side \\
    Line spacing & Font size $+\,1$\,pt (10\,pt at the selected size) \\
    Colors & Black text on a white background \\
    Wrapping & Text wrapped and paginated within the margins \\
    Resize alignment & Both image dimensions aligned to multiples of 32 \\
    Image pixel limits & Minimum 65,536; maximum 525,312 \\
    Patch and spatial merging & $16\times16$ pixels; $2\times2$ merging \\
    Processed 72-DPI page & $608\times832$ pixels; 494 visual tokens \\
    \bottomrule
  \end{tabular}
\end{table}

\subsection{Transcription Protocol and Readability Criterion}
\label{app:rendering-protocol}
\label{app:rendering-cer}

We use Qwen3.5-9B~\citep{qwen2026qwen35} on vLLM
0.19.1~\citep{kwon2023pagedattention} with deterministic decoding, thinking
disabled, and an 8,192-token context limit. A fixed prompt requests exact
transcription, preserving case, digits, punctuation, and nonsense strings.
The size sweep covers 15 fonts and eight sizes
($\{4,5,6,7,8,9,11,13\}$\,pt), sharing 30 random passages across all pairs
(3,600 images). Each passage contains 60 lowercase strings of lengths
sampled from $\{3,4,4,5,5,6,6,7,8\}$, with 6\% of characters replaced by
digits and another 6\% capitalized (seed 23).

Character error rate (CER) is
\begin{equation}
  c_i=\frac{\operatorname{Lev}(\mathcal N(\hat g_i),\mathcal N(g_i))}
  {\max\{|\mathcal N(g_i)|,1\}},
  \label{eq:rendering-cer}
\end{equation}
where $g_i$ and $\hat g_i$ are the reference and transcription,
$\mathcal N$ collapses repeated whitespace, and $\operatorname{Lev}$ is
Levenshtein distance~\citep{wagner1974string}. The sweep averages passage
CERs capped at one, averaging duplicate records, excluding failed requests,
and retaining truncated generations; error bars show one standard error.

We select the smallest tested size with mean CER $\leq5\%$: 9\,pt for
both DejaVu Sans and Verdana. Thresholds in
Table~\ref{tab:rendering-all-fonts} are interpolated linearly in log point
size, with 95\% intervals from 3,000 paired passage bootstrap resamples
(seed 1)~\citep{efron1979bootstrap}.

\subsection{Font Comparison and Visual-Token Cost}
\label{app:rendering-cost}

At 9\,pt, we measure uncapped CER on 400 shared passages (151,220 normalized
reference characters per font) and cost on 50 English passages assembled
from LongBench~\citep{bai2024longbench}, targeting 32,768 Qwen3.5 text tokens
(mean re-encoded length 32,762.8). Using the layout in
Table~\ref{tab:rendering-parameters}, mean visual-token cost is
\begin{equation}
  \overline V_f=\frac{1}{50}\sum_{i=1}^{50}\sum_{j=1}^{P_{if}}
  \frac{w_{ij}h_{ij}}{1024}.
  \label{eq:font-cost}
\end{equation}
Here $P_{if}$ is the page count and $w_{ij},h_{ij}$ are resized dimensions;
the denominator accounts for $16\times16$ patches with $2\times2$ merging.
Each 72-DPI A4 page costs 494 visual tokens, so cost depends on font-specific pagination,
including the last page, and excludes instructions, generation, and enhanced
regions.

\begin{table}[!htb]
  \centering
  \caption{All font candidates at 72 DPI, sorted by visual-token cost.
  CER uses the 400-passage 9\,pt set; threshold sizes and 95\% intervals
  use the separate 30-passage sweep. Pagination and visual tokens are
  means over the same 50 long passages. Bold identifies the selected font.}
  \label{tab:rendering-all-fonts}
  \small
  \setlength{\tabcolsep}{3pt}
  \begin{tabular}{llrrrr}
    \toprule
    Font & Family & CER (\%) & Threshold pt [95\% CI] & Pages & Visual tokens \\
    \midrule
    Liberation Sans Narrow & Sans & 17.53 & 11.69 [11.41, 11.91] & 12.18 & 6,016.9 \\
    Times New Roman & Serif & 13.15 & 10.87 [10.76, 11.01] & 13.54 & 6,688.8 \\
    Liberation Serif & Serif & 12.94 & 10.67 [10.56, 10.80] & 13.54 & 6,688.8 \\
    FreeSans & Sans & 9.30 & 10.60 [10.44, 10.77] & 14.50 & 7,163.0 \\
    Georgia & Serif & 10.76 & 10.73 [10.54, 10.90] & 14.82 & 7,321.1 \\
    Arial & Sans & 8.39 & 10.39 [10.19, 10.59] & 14.86 & 7,340.8 \\
    Tahoma & Sans & 7.31 & 10.02 [9.81, 10.21] & 14.94 & 7,380.4 \\
    Trebuchet MS & Sans & 6.04 & 9.50 [9.17, 9.81] & 15.20 & 7,508.8 \\
    \textbf{DejaVu Sans} & \textbf{Sans} & \textbf{3.91} & \textbf{8.69 [8.57, 8.81]} & \textbf{16.94} & \textbf{8,368.4} \\
    Verdana & Sans & 3.85 & 8.68 [8.53, 8.82] & 17.04 & 8,417.8 \\
    DejaVu Serif & Serif & 6.06 & 9.53 [9.26, 9.78] & 17.28 & 8,536.3 \\
    Consolas & Mono & 4.90 & 8.98 [8.90, 9.15] & 18.40 & 9,089.6 \\
    Courier New & Mono & 5.61 & 9.41 [9.04, 10.18] & 20.06 & 9,909.6 \\
    DejaVu Sans Mono & Mono & 2.70 & 8.40 [8.29, 8.49] & 20.06 & 9,909.6 \\
    FreeMono & Mono & 5.55 & 9.48 [9.15, 10.03] & 20.06 & 9,909.6 \\
    \bottomrule
  \end{tabular}
\end{table}

DejaVu Sans has the lowest measured cost among fonts meeting the 5\% CER criterion
(Table~\ref{tab:rendering-all-fonts}). Its random-text CER is close to
Verdana's (3.906\% versus 3.854\%; paired difference 95\% CI
$[-0.07,+0.18]$ percentage points, 20,000 bootstrap resamples, seed 0).
The main-text scatter includes only fonts with CER $\leq7\%$.

\subsection{Confusable Characters and Downstream Validation}
\label{app:rendering-confusables}
\label{app:rendering-controls}

The confusable-character test uses 300 passages shared by DejaVu Sans,
Verdana, Trebuchet MS, DejaVu Sans Mono, and Tahoma (116,925 normalized
characters per font; seed 31), enriched with sequences such as \texttt{cl},
\texttt{rn}, \texttt{vv}, and \texttt{il}. A unit is erroneous if any
of its reference positions overlaps a non-equal Levenshtein edit block;
overlapping occurrences count separately. Unit error rates pool erroneous
over total occurrences.
Figure~\ref{fig:font-selection}(c) reports DejaVu Sans minus Verdana with
95\% intervals from 8,000 paired passage bootstrap resamples (seed 0).
Overall CER is 0.30 percentage points lower for DejaVu Sans (95\% CI
$[-0.52,-0.09]$, 20,000 paired resamples).

A control uses 30 news/report passages in both fonts at eight sizes.
At 5\,pt, CER is 1.79\%/2.05\% for DejaVu Sans/Verdana versus
64.70\%/70.91\% on random text, showing how predictable language can mask
visual ambiguity and motivating random-text selection~\citep{gao2026decoupling}.

Under matched settings, \model with DejaVu Sans improves LongBench from
54.93 to 56.40 (Table~\ref{tab:training-stages}) and matches or exceeds
Verdana on every RULER v1 task, with the largest gains on
\texttt{single\_3} and \texttt{multikey\_3} (Table~\ref{tab:ruler-font}).
For example, in \texttt{single\_3}, the key \texttt{vague-ecology} maps to
the UUID \texttt{c6a7ee39-c4b0-42cc-97c5-24a55304317f}.
These arbitrary identifiers offer little semantic redundancy, so a single
misread hexadecimal digit can invalidate the answer.

\begin{table}[!htb]
\centering
\caption{FocusVTC font comparison on RULER v1 at 72 DPI (scores in \%).}
\label{tab:ruler-font}
\begingroup
\fontsize{8}{9.6}\selectfont
\setlength{\tabcolsep}{3pt}
\renewcommand{\arraystretch}{1.18}
\arrayrulecolor{ResultNavy}
\resizebox{\linewidth}{!}{%
\begin{tabular}{l*{10}{c}}
\rowcolor{ResultTableHeader}
\textcolor{white}{\textbf{Model / font}} & \textcolor{white}{\textbf{Single-1}} & \textcolor{white}{\textbf{Single-2}} & \textcolor{white}{\textbf{Single-3}} & \textcolor{white}{\textbf{MKey-1}} & \textcolor{white}{\textbf{MKey-2}} & \textcolor{white}{\textbf{MKey-3}} & \textcolor{white}{\textbf{MValue}} & \textcolor{white}{\textbf{MQuery}} & \textcolor{white}{\textbf{QA-1}} & \textcolor{white}{\textbf{QA-2}} \\
\rowcolor{ResultFocusStrong}
\textcolor{ResultTeal}{\textbf{FocusVTC}} (Ours) & 98.00 & 100.00 & 72.00 & 97.00 & 96.00 & 46.00 & 96.25 & 98.50 & 89.00 & 81.00 \\
\rowcolor{ResultFocus}
\textcolor{ResultTeal}{FocusVTC (Verdana)} (Ours) & 98.00 & 99.00 & 63.00 & 94.00 & 96.00 & 40.00 & 95.80 & 98.50 & 87.00 & 80.00 \\
\bottomrule
\end{tabular}}
\endgroup
\end{table}

\section{Data and Implementation Details}
\label{app:training-details}

REL-SFT and GRPO use AdamW~\citep{loshchilov2019adamw}, bf16 mixed
precision, gradient checkpointing, and gradient clipping at $1.0$.

\subsection{Training Data Details}
\label{app:training-data}

\begin{figure}[!htb]
  \centering
  \includegraphics[width=\linewidth]{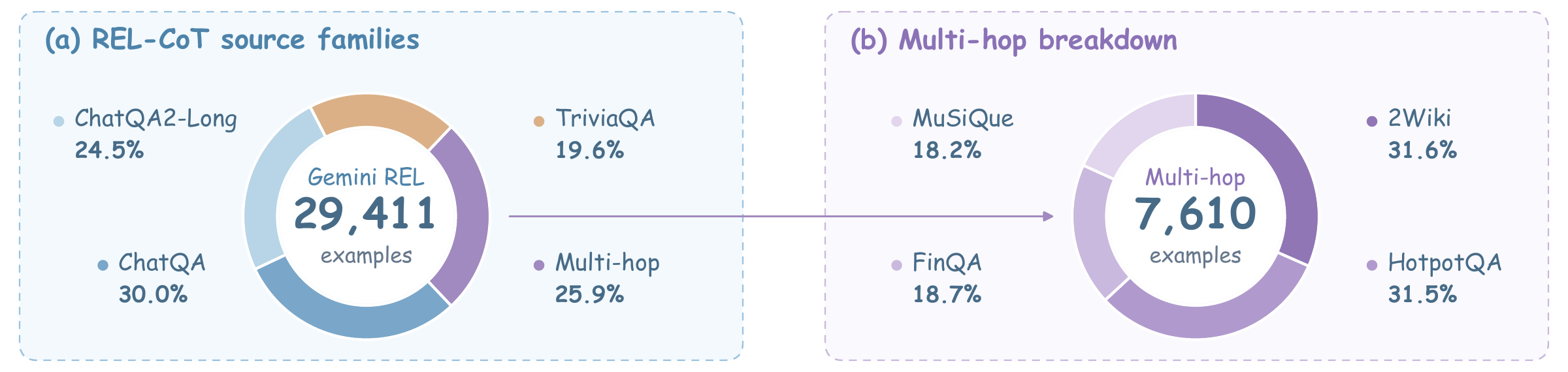}
  \caption{REL-CoT source families (left) and the Multi-hop breakdown
  (right), before resolution expansion. Percentages use each panel's total;
  the right panel combines long and original variants of each dataset.}
  \label{fig:rel-source-inventory}
\end{figure}

The 29,411 REL-CoT examples (Figure~\ref{fig:rel-source-inventory}) yield
205,877 candidate SFT samples after resolution expansion and before length
filtering. The ChatQA portion comprises DROP (3,327), Quoref (870),
ROPES (1,705), and TAT-QA (2,931). ChatQA2 contributes
NarrativeQA-131072 (2,018) and Long-SFT (5,195), while TriviaQA reading
comprehension and FinQA contribute 5,755 and 1,423 examples, respectively.
The multi-hop subset contains 285/2,123 examples from 2WikiMultihopQA,
382/2,014 from HotpotQA, and 277/1,106 from MuSiQue, where each pair denotes
original/long-context variants. Long multi-hop contexts append and shuffle
passages from the same dataset; TriviaQA concatenates retrieved passages,
and FinQA retains text and tables.

The GRPO set contains 10,000 prompts selected by the filtering procedure in
Section~\ref{sec:grpo}: 7,000 at 72 DPI, 2,000 at 96 DPI, and 1,000 at 144 DPI.

\subsection{Supervised Fine-Tuning Details}
\label{app:training-hyperparameters}
\label{app:sft-packing}

REL-SFT initializes \model from Qwen3.5-9B~\citep{qwen2026qwen35},
using the settings in Table~\ref{tab:sft-training-hyperparameters}.
Batch size counts packed sequences, and sequence length is measured in tokens.

\begin{table}[H]
  \centering
  \caption{\textbf{Stage 1: REL-SFT training hyperparameters.}}
  \label{tab:sft-training-hyperparameters}
  \begingroup
  \small
  \setlength{\tabcolsep}{3pt}
  \renewcommand{\arraystretch}{1.12}
  \arrayrulecolor{black}
  \begin{tabular*}{\linewidth}{@{\extracolsep{\fill}}*{8}{c}@{}}
    \toprule
    \shortstack{Global\\batch size} &
    \shortstack{Training\\steps} &
    \shortstack{Learning\\rate} &
    \shortstack{Weight\\decay} &
    \shortstack{LR\\schedule} &
    \shortstack{Warm-up\\steps} &
    \shortstack{Visual\\encoder} &
    \shortstack{Max. packed\\seq. length} \\
    \midrule
    8 & 7,000 & $10^{-6}$ & 0.01 & Cosine & 325 & Frozen & 32,768 \\
    \bottomrule
  \end{tabular*}
  \endgroup
\end{table}

\begin{figure}[!htb]
  \centering
  \begin{tikzpicture}[x=0.91cm,y=0.78cm,font=\small]
    \node[anchor=west] at (0,2.15) {Pool: $\leq256$ rows};
    \node[draw,rounded corners=2pt,fill=black!4,minimum width=2.55cm,
          minimum height=0.64cm] at (1.45,1.28) {Length $\leq 32{,}768$};
    \draw[-{Stealth[length=2mm]},thick] (3,1.28) -- (4,1.28);
    \node[align=center] at (3.5,0.52) {longest\\that fits};
    \draw[draw=black!60,fill=black!5] (4.2,0.85) rectangle (7.4,1.7);
    \draw[draw=black!60,fill=black!5] (7.4,0.85) rectangle (10,1.7);
    \draw[draw=black!60,fill=black!5] (10,0.85) rectangle (12.35,1.7);
    \fill[teal!25] (6.2,0.86) rectangle (7.39,1.69);
    \fill[teal!25] (9.1,0.86) rectangle (9.99,1.69);
    \fill[teal!25] (11.5,0.86) rectangle (12.34,1.69);
    \node at (5.8,1.29) {$x_1$};
    \node at (8.7,1.29) {$x_2$};
    \node at (11.15,1.29) {$x_3$};
    \node[anchor=west] at (4.2,2.15) {One pack: $\sum_i \ell_i\leq32{,}768$};
    \draw[densely dashed,black!60] (7.4,0.35) -- (7.4,1.95);
    \draw[densely dashed,black!60] (10,0.35) -- (10,1.95);
    \node[align=center] at (8.25,-0.13)
      {Each boundary resets attention, recurrent state, and positions};
    \fill[black!8,draw=black!40] (1,-0.83) rectangle (1.25,-0.58);
    \node[anchor=west] at (1.35,-0.7) {Conditioning span};
    \fill[teal!25,draw=black!40] (6,-0.83) rectangle (6.25,-0.58);
    \node[anchor=west] at (6.35,-0.7) {Target span};
  \end{tikzpicture}
  \caption{SFT sequence packing. Gray spans denote conditioning inputs;
  teal spans denote assistant targets.}
  \label{fig:sft-packing}
\end{figure}
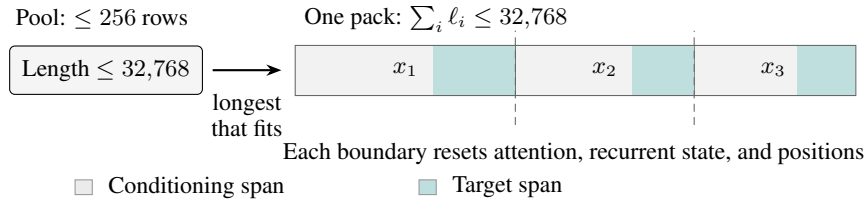

We pack complete examples into 32,768-token sequences
(Figure~\ref{fig:sft-packing}). Lengths are measured after chat templating
and image preprocessing, including assistant targets; longer examples are
excluded. Each worker selects the longest fitting sample from a pool of
up to 256, refilling below 128 samples or before closing a pack with no
fitting candidate. SFT uses FSDP2, FlashAttention-2, and fused AdamW,
achieving 99.85\% packing occupancy and approximately 51\% model FLOPs
utilization (MFU). The near-full packs minimize unused sequence capacity,
while the observed MFU reflects efficient use of the training hardware.
Training takes 26.7 hours on eight GPUs.

\subsection{Group Relative Policy Optimization Details}
\label{app:rollout-protocol}
\label{app:tool-protocol}

GRPO starts from the REL-SFT checkpoint with the settings in
Table~\ref{tab:grpo-training-hyperparameters}. The direct-GRPO baseline
FocusVTC w/o SFT also trains for 150 updates.
Global and PPO mini-batch sizes count prompts before rollout expansion;
each update samples $64\times8=512$ trajectories. The PPO micro-batch
counts trajectories per GPU. All prompt, response, and generation lengths
are in tokens ($1\mathrm{K}=1{,}024$).

\begin{table}[H]
  \centering
  \caption{\textbf{Stage 2: GRPO training hyperparameters.}}
  \label{tab:grpo-training-hyperparameters}
  \begingroup
  \footnotesize
  \setlength{\tabcolsep}{1.5pt}
  \renewcommand{\arraystretch}{1.12}
  \arrayrulecolor{black}
  \resizebox{\linewidth}{!}{%
  \begin{tabular}{@{}*{16}{c}@{}}
    \toprule
    \shortstack{Global\\batch} &
    \shortstack{Train\\steps} &
    LR &
    \shortstack{Weight\\decay} &
    \shortstack{LR\\schedule} &
    \shortstack{Warmup\\steps} &
    \shortstack{Visual\\encoder} &
    \shortstack{Max.\\prompt /\\response} &
    \shortstack{Max. gen.\\/ turn} &
    \shortstack{Rollouts /\\prompt} &
    \shortstack{PPO\\mini-batch} &
    \shortstack{PPO micro-\\batch /\\GPU} &
    \shortstack{Temp. /\\top-$p$} &
    \shortstack{Max.\\turns /\\tool calls} &
    \shortstack{$g$ exponent\\$\gamma$} &
    \shortstack{Clip\\$\epsilon$} \\
    \midrule
    64 & 150 & $10^{-6}$ & 0.01 & Constant & 0 & Trainable &
    8K / 10K & 2K & 8 & 64 & 1 & 1.0 / 1.0 & 9 / 8 & 2 & 0.2 \\
    \bottomrule
  \end{tabular}}
  \endgroup
\end{table}

Online rollouts are sampled independently of the data-filtering trajectories
in Section~\ref{sec:grpo}.
The post-prompt budget includes tool observations. Rollouts end with an
\texttt{<answer>...</answer>} block or an exhausted length or interaction
budget. Training takes approximately 139.2 hours on eight GPUs.

The policy uses \texttt{Enhance\_Region(page, bbox)} to request regions from
aligned 144-DPI pages. Page indices are one-based, and bounding boxes satisfy
$0\leq x_1<x_2\leq1000$ and $0\leq y_1<y_2\leq1000$.
Malformed or unexecutable calls consume an attempt and return an error.
Table~\ref{tab:reward-boundaries} summarizes reward behavior for representative
interaction cases.

\begin{table}[!htb]
  \centering
  \caption{Representative cases under the reward function. Unless otherwise
  stated, terminal answer formatting is valid.}
  \label{tab:reward-boundaries}
  \small
  \begin{tabular}{@{}p{0.35\linewidth}p{0.59\linewidth}@{}}
    \toprule
    Situation & Consequence \\
    \midrule
    Fully correct answer, no tool calls & $R_{\mathrm{acc}}=R_{\mathrm{fmt}}=1$ and $R_{\mathrm{tool}}=0$, giving $R=1$. \\
    Answer not fully correct & The tool-bonus term is gated off. \\
    Full-page request & $s_i=1$: the area factor is zero, so this call contributes zero to the matching sum. \\
    Repeated requests for the same evidence & Each annotated evidence region can be matched at most once. \\
    Excess or malformed tool calls & Both set $R_{\mathrm{fmt}}=0$. For $N>E$, $E_{\mathrm{call}}=E/N<1$; malformed calls count in $N$ but are excluded from matching. \\
    Initial view at 144 DPI & $g(144)=0$, giving $R=0.8R_{\mathrm{acc}}+0.2R_{\mathrm{fmt}}$. \\
    \bottomrule
  \end{tabular}
\end{table}

\section{Evaluation Results Detail And Efficiency}
\label{app:evaluation-details}
\label{app:eval-protocol}

Unless otherwise specified, we evaluate Qwen3.5-9B (with and without tools)
and FocusVTC in thinking mode using the official Qwen3.5 sampling
settings~\citep{qwen2026qwen35}: temperature $=1.0$, top-$p=0.95$,
top-$k=20$, min-$p=0.0$, presence penalty $=1.5$, and repetition penalty
$=1.0$.
Tool-enabled evaluation allows up to eight calls over nine rounds, with
an 8,192-token limit per generation and a 20,480-token post-prompt
trajectory limit, including tool observations.
The training examples used for REL-SFT and GRPO do not overlap with the
test examples in any of our evaluation benchmarks.

\subsection{Token-Cost Accounting}
\label{app:cost-accounting}

\paragraph{Prompt and extra observations.}
Let $P$ be the number of tokens in the initial prompt and $O$ the total
number of extra observation tokens appended during tool use, including
returned images and environment text. For RULER, MRCR, and LongBench, we define input
compression as
\begin{equation}
  C_{\mathrm{context}}=\frac{T_{\mathrm{text}}}{P+O},
\end{equation}
where $T_{\mathrm{text}}$ is the token count of the text-context reference.
Each extra observation is counted once;
Table~\ref{tab:ruler-cost-accounting} reports token costs for
RULER and LongBench, including the RULER decomposition underlying
Figure~\ref{fig:ruler-dpi}.

\begin{table}[!htb]
  \centering
  \caption{Token costs and compression. $P$ and $O$ denote initial prompt
  and extra-observation tokens.
  Text references are 8,400/9,201/11,313 tokens for RULER v1/v2 and LongBench.}
  \label{tab:ruler-cost-accounting}
  \small
  \setlength{\tabcolsep}{4pt}
  \begin{tabular}{@{}l c c c c c c c@{}}
    \toprule
    & & \multicolumn{2}{c}{Glyph \citep{cheng2025glyph}} & \multicolumn{4}{c}{FocusVTC (Ours)} \\
    \cmidrule(lr){3-4}\cmidrule(l){5-8}
    Benchmark & DPI & P & Text/P & $P$ & $O$ & $P+O$ & Text/$(P+O)$ \\
    \midrule
    RULER v1 & 48 & 1,491 & 5.6$\times$ & 1,218 & 1,768 & 2,986 & 2.8$\times$ \\
     & 60 & 2,061 & 4.1$\times$ & 1,655 & 1,392 & 3,047 & 2.8$\times$ \\
    \rowcolor{ResultFocus}
     & 72 & 2,765 & 3.0$\times$ & 2,154 & 723 & 2,877 & 2.9$\times$ \\
     & 84 & 3,517 & 2.4$\times$ & 2,729 & 698 & 3,427 & 2.5$\times$ \\
     & 96 & 4,522 & 1.9$\times$ & 3,544 & 681 & 4,225 & 2.0$\times$ \\
     & 120 & 6,731 & 1.2$\times$ & 5,277 & 634 & 5,911 & 1.4$\times$ \\
     & 144 & 9,153 & 0.9$\times$ & 7,290 & 625 & 7,915 & 1.1$\times$ \\
    \midrule
    RULER v2 & 48 & 2,209 & 4.2$\times$ & 1,495 & 1,861 & 3,356 & 2.7$\times$ \\
     & 60 & 2,793 & 3.3$\times$ & 1,902 & 1,458 & 3,360 & 2.7$\times$ \\
    \rowcolor{ResultFocus}
     & 72 & 3,440 & 2.7$\times$ & 2,399 & 852 & 3,251 & 2.8$\times$ \\
     & 84 & 4,138 & 2.2$\times$ & 3,050 & 834 & 3,884 & 2.4$\times$ \\
     & 96 & 5,032 & 1.8$\times$ & 3,731 & 803 & 4,534 & 2.0$\times$ \\
     & 120 & 7,291 & 1.3$\times$ & 5,450 & 818 & 6,268 & 1.5$\times$ \\
     & 144 & 10,052 & 0.9$\times$ & 7,585 & 834 & 8,419 & 1.1$\times$ \\
    \midrule
    LongBench & 72 & 3,691 & 3.1$\times$ & 3,014 & 723 & 3,737 & 3.0$\times$ \\
    \bottomrule
  \end{tabular}
\end{table}

At 72 DPI, extra observations account for 25.1\%/26.2\% of
RULER v1/v2 input tokens, giving compression of
$2.9\times/2.8\times$. These ratios divide the mean text-reference length
by the mean $P+O$ length.
For MRCR's longest bin, the text reference is 197,909 tokens and the
prompt is 53,181 tokens. Adding the two/four/eight-needle observations
of 7,296/10,290/13,048 tokens gives $3.3\times/3.1\times/3.0\times$
compression.

\subsection{End-to-End Latency}
\label{app:online-latency}

We compare Qwen3.5-9B Text and FocusVTC on 300 paired MRCR four-needle
examples (64K--128K). Each system uses two NVIDIA A100 80GB GPUs (TP=2),
processing one request at a time. We time each example once after warmup,
using greedy decoding with thinking enabled and an 8,192-token cumulative
generation budget. Online latency is computed by subtracting each example's
page-rendering time from its recorded end-to-end latency, treating rendering
as offline preprocessing. It includes image loading, processing and encoding,
all inference rounds, and tool execution.

\begin{table}[H]
  \centering
  \caption{End-to-end latency on 300 paired MRCR four-needle
  examples with 64K--128K contexts. Values exclude recorded page-rendering
  time but include all online inference and tool rounds. Prefill timings are
  measured separately; tool-round prefill is cumulative per sample.
  Lower is better.}
  \label{tab:online-latency}
  \small
  \setlength{\tabcolsep}{8pt}
  \begin{tabular}{@{}c c c c@{}}
    \toprule
    Model & \shortstack{Latency \\(s)}
    & \shortstack{Initial prefill\\(s)}
    & \shortstack{Tool-round prefill\\(s/sample)} \\
    \midrule
    Text (Qwen3.5-9B) & 187.09 & 6.42 & -- \\
    \rowcolor{ResultFocus}
    FocusVTC (Ours) & \textbf{67.06} & \textbf{4.14} & \textbf{1.34} \\
    \bottomrule
  \end{tabular}
\end{table}

Table~\ref{tab:online-latency} shows that FocusVTC reduces mean online
latency from 187.09\,s to 67.06\,s, a 64.2\% reduction ($2.79\times$
speedup). Text's longer input is accompanied by substantially more
generated tokens: 5,411.30 per example versus 1,529.77 for
FocusVTC, including thinking, answers, and all tool rounds. These results
show that FocusVTC's compressed visual input is accompanied by fewer
generated tokens and lower online latency.

\subsection{Detailed Long-Context Results}
\label{app:long-context-results}
\label{app:training-stages}
\label{app:mrcr-results}

Tables~\ref{tab:longbench13-stages}, \ref{tab:vtcbench-stages}, and~\ref{tab:mrcr}
report detailed LongBench, VTCBench, and MRCR results for the training and
tool-use variants.
VTCBench~\citep{zhao2025vtcbench} is designed specifically to evaluate
long-context understanding under visual text compression.
Retrieval focuses on locating and aggregating information, Reasoning on
inferring associations beyond direct lexical matching, and Memory on
recalling and using information from long dialogue histories.
On LongBench, the gap between English and Chinese passage retrieval narrows
from 50.27 points for FocusVTC w/o GRPO to 0.50 for FocusVTC, whose final
scores are 98.50 and 98.00.
On MRCR, FocusVTC's scores decrease with needle count in every context-length
bin; even at 4K--8K, the scores are 82.25, 65.89, and 50.84 for two, four,
and eight needles, respectively.

\begingroup
\raggedbottom
\begin{table}[!htb]
\caption{Detailed LongBench scores (\%) for training-stage variants. Avg is the unweighted arithmetic mean of the 11 displayed non-code task scores. The final FocusVTC row is included as reference. Vision denotes 72-DPI pages. +tools uses tool access without additional training.}
\label{tab:longbench13-stages}
\centering
\begingroup
\fontsize{8}{9.6}\selectfont
\setlength{\tabcolsep}{2.3pt}
\renewcommand{\arraystretch}{1.18}
\arrayrulecolor{ResultNavy}
\resizebox{\linewidth}{!}{%
\begin{tabular}{lc*{12}{c}}
\rowcolor{ResultTableHeader}
\textcolor{white}{\textbf{Model / stage}} & \textcolor{white}{\textbf{Input}} & \multicolumn{3}{c}{\textcolor{white}{\textbf{Single-doc QA}}} & \multicolumn{2}{c}{\textcolor{white}{\textbf{Multi-doc QA}}} & \multicolumn{2}{c}{\textcolor{white}{\textbf{Summarization}}} & \multicolumn{2}{c}{\textcolor{white}{\textbf{Few-shot}}} & \multicolumn{2}{c}{\textcolor{white}{\textbf{Synthetic}}} & \textcolor{white}{\textbf{Overall}} \\
\rowcolor{ResultHeader}
& & QP & MF-En & MF-Zh & DuR & 2Wiki & MNews & QMSum & SAMSum & Trivia & PR-En & PR-Zh & \textbf{Avg} \\
\addlinespace[2.5pt]
\rowcolor{ResultFocus}
\textcolor{ResultTeal}{\textbf{FocusVTC w/o SFT}} (Ours) & Vision & 43.77 & \underline{46.73} & 53.80 & 26.74 & \underline{61.26} & \textbf{24.07} & \textbf{23.95} & 31.56 & 85.31 & 72.32 & 72.74 & \cellcolor{ResultTeal!15}49.30 \\
\rowcolor{ResultFocus}
\textcolor{ResultTeal}{\textbf{FocusVTC w/o GRPO}} (Ours) & Vision & 34.52 & 39.94 & 31.77 & 14.73 & 56.54 & 17.47 & 16.69 & 31.64 & 89.23 & 67.12 & 16.85 & \cellcolor{ResultTeal!15}37.86 \\
\rowcolor{ResultFocus}
\textcolor{ResultTeal}{\textbf{FocusVTC w/o GRPO+tools}} (Ours) & Vision & 16.46 & 32.70 & 23.16 & 8.42 & 49.70 & 2.96 & 3.79 & 3.88 & 56.42 & 65.19 & 16.83 & \cellcolor{ResultTeal!15}25.41 \\
\rowcolor{ResultFocus}
\textcolor{ResultTeal}{\textbf{FocusVTC-50}} (Ours) & Vision & \underline{47.61} & 41.73 & 49.21 & 29.80 & 52.73 & 21.47 & 12.15 & 30.78 & 87.21 & 94.05 & 93.66 & \cellcolor{ResultTeal!15}50.95 \\
\rowcolor{ResultFocus}
\textcolor{ResultTeal}{\textbf{FocusVTC-100}} (Ours) & Vision & 46.90 & 45.06 & \underline{54.22} & \underline{30.10} & 61.14 & 22.47 & 20.06 & \underline{32.76} & \underline{90.10} & \underline{95.73} & \underline{96.00} & \cellcolor{ResultTeal!15}\underline{54.05} \\
\rowcolor{ResultFocus}
\textcolor{ResultTeal}{\textbf{FocusVTC w/o tools}} (Ours) & Vision & 34.02 & 41.38 & 21.70 & 13.26 & 59.62 & 17.47 & 12.36 & 9.10 & 81.83 & 79.00 & 35.26 & \cellcolor{ResultTeal!15}36.82 \\
\rowcolor{ResultFocusStrong}
\textcolor{ResultTeal}{\textbf{FocusVTC}} (Ours) & Vision & \textbf{47.71} & \textbf{48.95} & \textbf{57.78} & \textbf{30.21} & \textbf{69.04} & \underline{23.57} & \underline{21.71} & \textbf{33.87} & \textbf{91.01} & \textbf{98.50} & \textbf{98.00} & \cellcolor{ResultTeal!24}\textbf{56.40} \\
\specialrule{0.35pt}{3pt}{3pt}
\rowcolor{ResultStripe}
\shortstack[l]{Qwen3.5-9B+tools\\\citep{qwen2026qwen35}} & Vision & 15.58 & 34.22 & 27.10 & 22.03 & 47.81 & 3.16 & 4.51 & 4.88 & 61.70 & 66.01 & 12.40 & \cellcolor{ResultHeader}27.22 \\
\bottomrule
\end{tabular}%
}
\endgroup
\end{table}

\begin{table}[!htb]
\centering
\caption{Detailed VTCBench results (\%) for training-stage variants. Length labels are bin upper bounds (K tokens). Each Avg is the unweighted arithmetic mean of the four displayed length-bin scores. The final FocusVTC row is included as reference. Vision denotes 72-DPI pages. +tools uses tool access without additional training.}
\label{tab:vtcbench-stages}
\begingroup
\fontsize{8}{9.6}\selectfont
\setlength{\tabcolsep}{2.3pt}
\renewcommand{\arraystretch}{1.18}
\arrayrulecolor{ResultNavy}
\resizebox{\linewidth}{!}{%
\begin{tabular}{lcccccc@{\hspace{6pt}}ccccc@{\hspace{6pt}}ccccc}
\rowcolor{ResultTableHeader}
\textcolor{white}{\textbf{Model / stage}} & \textcolor{white}{\textbf{Input}} & \multicolumn{5}{c}{\textcolor{white}{\textbf{Retrieval}}} & \multicolumn{5}{c}{\textcolor{white}{\textbf{Reasoning}}} & \multicolumn{5}{c}{\textcolor{white}{\textbf{Memory}}} \\
\rowcolor{ResultHeader}
 & & 8 & 16 & 32 & 64 & \textbf{Avg} & 8 & 16 & 32 & 64 & \textbf{Avg} & 8 & 16 & 32 & 64 & \textbf{Avg} \\
\addlinespace[2.5pt]
\rowcolor{ResultFocus}
\textcolor{ResultTeal}{\textbf{FocusVTC w/o SFT}} (Ours) & Vision & \textbf{99.83} & 85.55 & \underline{82.22} & \underline{81.07} & \cellcolor{ResultTeal!15}\underline{87.17} & \textbf{50.88} & \textbf{49.09} & \underline{33.96} & \underline{18.65} & \cellcolor{ResultTeal!15}\textbf{38.15} & \underline{31.24} & \textbf{22.21} & 21.57 & 19.20 & \cellcolor{ResultTeal!15}\underline{23.56} \\
\rowcolor{ResultFocus}
\textcolor{ResultTeal}{\textbf{FocusVTC w/o GRPO}} (Ours) & Vision & 87.78 & 79.56 & 80.21 & 73.52 & \cellcolor{ResultTeal!15}80.27 & 40.17 & 25.68 & 12.04 & 3.57 & \cellcolor{ResultTeal!15}20.37 & 12.12 & 17.98 & 20.60 & 20.00 & \cellcolor{ResultTeal!15}17.68 \\
\rowcolor{ResultFocus}
\textcolor{ResultTeal}{\textbf{FocusVTC w/o GRPO+tools}} (Ours) & Vision & 66.28 & 59.67 & 51.72 & 39.34 & \cellcolor{ResultTeal!15}54.25 & 7.82 & 1.35 & 0.93 & 0.00 & \cellcolor{ResultTeal!15}2.53 & 7.50 & 4.23 & 6.18 & 9.79 & \cellcolor{ResultTeal!15}6.93 \\
\rowcolor{ResultFocus}
\textcolor{ResultTeal}{\textbf{FocusVTC-50}} (Ours) & Vision & 96.96 & 80.65 & 81.97 & 65.56 & \cellcolor{ResultTeal!15}81.29 & 38.51 & 29.37 & 14.22 & 7.00 & \cellcolor{ResultTeal!15}22.28 & 17.63 & 13.65 & 18.06 & \underline{24.99} & \cellcolor{ResultTeal!15}18.58 \\
\rowcolor{ResultFocus}
\textcolor{ResultTeal}{\textbf{FocusVTC-100}} (Ours) & Vision & 97.94 & \underline{87.85} & \textbf{85.34} & 77.05 & \cellcolor{ResultTeal!15}87.05 & 39.53 & 29.73 & 16.67 & 11.21 & \cellcolor{ResultTeal!15}24.29 & 25.53 & 17.84 & 20.85 & \textbf{25.64} & \cellcolor{ResultTeal!15}22.47 \\
\rowcolor{ResultFocus}
\textcolor{ResultTeal}{\textbf{FocusVTC w/o tools}} (Ours) & Vision & 72.94 & 75.14 & 58.62 & 62.30 & \cellcolor{ResultTeal!15}67.25 & 10.15 & 1.35 & 0.93 & 0.00 & \cellcolor{ResultTeal!15}3.11 & 26.27 & 20.50 & \textbf{24.88} & 20.15 & \cellcolor{ResultTeal!15}22.95 \\
\rowcolor{ResultFocusStrong}
\textcolor{ResultTeal}{\textbf{FocusVTC}} (Ours) & Vision & \underline{98.39} & \textbf{93.37} & \textbf{85.34} & \textbf{86.89} & \cellcolor{ResultTeal!24}\textbf{91.00} & \underline{45.24} & \underline{43.24} & \textbf{35.19} & \textbf{21.33} & \cellcolor{ResultTeal!24}\underline{36.25} & \textbf{32.98} & \underline{22.07} & \underline{24.64} & \textbf{25.64} & \cellcolor{ResultTeal!24}\textbf{26.33} \\
\specialrule{0.35pt}{3pt}{3pt}
\rowcolor{ResultStripe}
\shortstack[l]{Qwen3.5-9B+tools\\\citep{qwen2026qwen35}} & Vision & 58.26 & 59.12 & 45.69 & 34.43 & \cellcolor{ResultHeader}49.38 & 7.19 & 1.35 & 2.78 & 0.00 & \cellcolor{ResultHeader}2.83 & 3.08 & 2.56 & 5.80 & 1.77 & \cellcolor{ResultHeader}3.30 \\
\bottomrule
\end{tabular}%
}
\endgroup
\end{table}

\begin{table}[H]
\caption{MRCR performance (\%) for two, four, and eight needles. Length labels are bin upper bounds (K tokens). Avg is the unweighted arithmetic mean of the six length-bin scores. Vision denotes 72-DPI pages. +tools uses tool access without additional training.}
\label{tab:mrcr}
\centering
\begingroup
\fontsize{8}{9.6}\selectfont
\setlength{\tabcolsep}{1.0pt}
\renewcommand{\arraystretch}{1.18}
\arrayrulecolor{ResultNavy}
\renewcommand{\arraystretch}{1.3}
\resizebox{\linewidth}{!}{%
\begin{tabular}{lc*{7}{c}@{\hspace{4pt}}*{7}{c}@{\hspace{4pt}}*{7}{c}}
\rowcolor{ResultTableHeader}
\textcolor{white}{\textbf{Model / stage}} & \textcolor{white}{\textbf{Input}} & \multicolumn{7}{c}{\textcolor{white}{\textbf{2 needles}}} & \multicolumn{7}{c}{\textcolor{white}{\textbf{4 needles}}} & \multicolumn{7}{c}{\textcolor{white}{\textbf{8 needles}}} \\
\rowcolor{ResultHeader}
 & & 8 & 16 & 32 & 64 & 128 & 256 & \textbf{Avg} & 8 & 16 & 32 & 64 & 128 & 256 & \textbf{Avg} & 8 & 16 & 32 & 64 & 128 & 256 & \textbf{Avg} \\
\shortstack[l]{LLaMA-3.1-8B\\\citep{grattafiori2024llama3}} & Text & 54.27 & \textbf{53.21} & \textbf{51.05} & 29.81 & 24.98 & 20.90 & \cellcolor{ResultHeader}39.04 & \underline{33.42} & \underline{25.97} & \underline{22.73} & \textbf{26.97} & 12.68 & 6.00 & \cellcolor{ResultHeader}21.30 & \textbf{23.80} & 17.69 & \underline{19.85} & \underline{17.72} & 11.79 & 7.80 & \cellcolor{ResultHeader}16.44 \\
\rowcolor{ResultStripe}
\shortstack[l]{Qwen2.5-7B\\\citep{yang2025qwen25million}} & Text & 45.92 & 51.07 & 46.97 & \underline{34.67} & \underline{37.57} & \underline{37.60} & \cellcolor{ResultHeader}\underline{42.30} & 25.96 & 20.13 & 19.93 & 24.25 & 17.29 & 12.30 & \cellcolor{ResultHeader}19.98 & 17.64 & 19.48 & 12.41 & 14.80 & 14.24 & \textbf{13.70} & \cellcolor{ResultHeader}15.38 \\
\shortstack[l]{Qwen3-8B\\\citep{yang2025qwen3}} & Text & \textbf{58.95} & 41.18 & 36.18 & 24.99 & 20.89 & 17.50 & \cellcolor{ResultHeader}33.28 & 29.34 & 22.67 & 20.34 & 23.63 & \underline{19.11} & \underline{15.50} & \cellcolor{ResultHeader}\underline{21.77} & 18.75 & \underline{19.69} & 16.81 & \textbf{17.86} & \underline{15.00} & \underline{12.60} & \cellcolor{ResultHeader}\underline{16.79} \\
\rowcolor{ResultStripe}
\shortstack[l]{GLM-4-9B\\\citep{glm2024chatglm}} & Text & 39.77 & 15.87 & 18.42 & 18.63 & 18.42 & 18.20 & \cellcolor{ResultHeader}21.55 & 15.17 & 13.78 & 9.18 & 20.27 & 15.05 & 11.20 & \cellcolor{ResultHeader}14.11 & 14.55 & 9.65 & 9.34 & 9.47 & 8.97 & 8.50 & \cellcolor{ResultHeader}10.08 \\
\shortstack[l]{Qwen3.5-9B\\\citep{qwen2026qwen35}} & Text & \underline{54.96} & \underline{51.59} & \underline{48.59} & \textbf{43.44} & \textbf{41.48} & \textbf{39.08} & \cellcolor{ResultHeader}\textbf{46.52} & \textbf{36.72} & \textbf{35.02} & \textbf{34.55} & \underline{25.44} & \textbf{25.82} & \textbf{21.66} & \cellcolor{ResultHeader}\textbf{29.87} & \underline{23.04} & \textbf{20.50} & \textbf{21.53} & 12.68 & \textbf{21.13} & 12.48 & \cellcolor{ResultHeader}\textbf{18.56} \\
\specialrule{0.45pt}{5pt}{5pt}
\rowcolor{ResultStripe}
\shortstack[l]{Glyph\\\citep{cheng2025glyph}} & Vision & 58.54 & 58.52 & 47.54 & 41.26 & 38.42 & 11.48 & \cellcolor{ResultHeader}42.63 & 34.56 & 39.26 & 33.71 & 21.96 & 19.43 & 6.92 & \cellcolor{ResultHeader}25.97 & 19.04 & 22.61 & 25.47 & 14.74 & 14.62 & 2.94 & \cellcolor{ResultHeader}16.57 \\
\addlinespace[2.5pt]
\rowcolor{ResultFocus}
\textcolor{ResultTeal}{\textbf{FocusVTC w/o SFT}} (Ours) & Vision & 74.77 & \underline{72.32} & 54.81 & 52.91 & \underline{46.37} & \underline{41.13} & \cellcolor{ResultTeal!15}\underline{57.05} & 51.37 & 46.15 & 47.32 & 34.17 & \underline{33.54} & \textbf{23.62} & \cellcolor{ResultTeal!15}39.36 & \underline{42.63} & 30.84 & 26.66 & 22.82 & \underline{21.80} & \underline{15.23} & \cellcolor{ResultTeal!15}\underline{26.66} \\
\rowcolor{ResultFocus}
\textcolor{ResultTeal}{\textbf{FocusVTC w/o GRPO}} (Ours) & Vision & 64.19 & 62.79 & 48.05 & 46.28 & 30.10 & 18.20 & \cellcolor{ResultTeal!15}44.94 & 34.64 & 42.73 & 30.00 & 33.57 & 20.43 & 11.58 & \cellcolor{ResultTeal!15}28.83 & 26.81 & 24.76 & \textbf{31.23} & 18.08 & 15.75 & 11.15 & \cellcolor{ResultTeal!15}21.30 \\
\rowcolor{ResultFocus}
\textcolor{ResultTeal}{\textbf{FocusVTC w/o GRPO+tools}} (Ours) & Vision & 41.36 & 14.77 & 23.75 & 25.02 & 15.11 & 13.48 & \cellcolor{ResultTeal!15}22.25 & 14.74 & 6.45 & 13.58 & 28.44 & 17.75 & 9.56 & \cellcolor{ResultTeal!15}15.09 & 10.47 & 8.83 & 13.48 & 11.28 & 8.93 & 0.77 & \cellcolor{ResultTeal!15}8.96 \\
\rowcolor{ResultFocus}
\textcolor{ResultTeal}{\textbf{FocusVTC-50}} (Ours) & Vision & 77.50 & 56.45 & 51.58 & 55.24 & 30.01 & 17.39 & \cellcolor{ResultTeal!15}48.03 & 62.00 & 43.12 & 42.40 & 31.19 & 20.29 & 14.22 & \cellcolor{ResultTeal!15}35.54 & 19.37 & 17.17 & 22.17 & 15.48 & 10.87 & 4.49 & \cellcolor{ResultTeal!15}14.93 \\
\rowcolor{ResultFocus}
\textcolor{ResultTeal}{\textbf{FocusVTC-100}} (Ours) & Vision & \underline{79.39} & 68.12 & 55.96 & 55.68 & 39.58 & 28.68 & \cellcolor{ResultTeal!15}54.57 & \underline{64.68} & 45.26 & \underline{48.48} & 33.59 & 30.67 & 19.41 & \cellcolor{ResultTeal!15}\underline{40.35} & 35.01 & 23.56 & 25.25 & 23.69 & 17.11 & 10.02 & \cellcolor{ResultTeal!15}22.44 \\
\rowcolor{ResultFocus}
\textcolor{ResultTeal}{\textbf{FocusVTC w/o tools}} (Ours) & Vision & 76.61 & 63.66 & \textbf{63.74} & \underline{56.04} & 34.51 & 17.45 & \cellcolor{ResultTeal!15}52.00 & 45.22 & \textbf{53.62} & 47.44 & \underline{37.06} & 25.87 & 16.70 & \cellcolor{ResultTeal!15}37.65 & 28.31 & \underline{32.47} & 22.67 & \underline{26.40} & 17.58 & 13.56 & \cellcolor{ResultTeal!15}23.50 \\
\rowcolor{ResultFocusStrong}
\textcolor{ResultTeal}{\textbf{FocusVTC}} (Ours) & Vision & \textbf{82.25} & \textbf{78.64} & \underline{56.58} & \textbf{56.98} & \textbf{47.55} & \textbf{42.56} & \cellcolor{ResultTeal!24}\textbf{60.76} & \textbf{65.89} & \underline{49.50} & \textbf{54.23} & \textbf{37.29} & \textbf{41.97} & \underline{22.36} & \cellcolor{ResultTeal!24}\textbf{45.21} & \textbf{50.84} & \textbf{33.65} & \underline{28.80} & \textbf{28.36} & \textbf{26.90} & \textbf{15.69} & \cellcolor{ResultTeal!24}\textbf{30.71} \\
\specialrule{0.35pt}{3pt}{3pt}
\rowcolor{ResultStripe}
\shortstack[l]{Qwen3.5-9B+tools\\\citep{qwen2026qwen35}} & Vision & 37.67 & 25.63 & 28.57 & 12.47 & 11.10 & 8.76 & \cellcolor{ResultHeader}20.70 & 24.32 & 17.54 & 14.56 & 13.29 & 8.93 & 7.86 & \cellcolor{ResultHeader}14.42 & 10.29 & 9.56 & 6.12 & 8.24 & 4.20 & 3.36 & \cellcolor{ResultHeader}6.96 \\
\bottomrule
\end{tabular}%
}
\endgroup
\end{table}

\subsection{Detailed RULER Results at 72 DPI}
\label{app:sft-tools-ruler-results}
\label{app:ruler-details}

Tables~\ref{tab:ruler-v1-details} and~\ref{tab:ruler-v2-details} report
the 72-DPI task-level results for the visual models and their training
and tool configurations. RULER v1 is the original RULER benchmark
~\citep{hsieh2024ruler}, evaluated here on ten retrieval and QA tasks,
excluding variable tracing and the two word-frequency tasks.
RULER v2
extends evaluation from retrieval to retrieval combined with reasoning:
it covers multi-key retrieval (MK-NIAH), multi-value retrieval (MV-NIAH),
and multi-document QA, each at four difficulty levels (Basic, Easy, Medium,
and Hard), yielding twelve task--difficulty combinations.
Each version's Avg is the unweighted mean of its subset scores.
For the two new measured variants, we recompute the v1 average over these
ten tasks, excluding the three additional tasks in their run summaries.
The retained v2 scorer permits relaxed matching on some tasks; neither
run receives additional credit from reasoning text, which was not saved
in the RULER predictions.

\begin{table}[H]
\centering
\caption{RULER v1 task scores (\%) at 72 DPI. Avg is the unweighted arithmetic mean of the 10 displayed task scores. Best and second-best scores are bold and underlined, respectively. Qwen3.5-9B+tools uses tool access without additional training.}
\label{tab:ruler-v1-details}
\label{tab:sft-tools-ruler}
\begingroup
\fontsize{8}{9.6}\selectfont
\setlength{\tabcolsep}{1.0pt}
\renewcommand{\arraystretch}{1.18}
\arrayrulecolor{ResultNavy}
\resizebox{\linewidth}{!}{%
\begin{tabular}{l*{11}{c}}
\rowcolor{ResultTableHeader}
\textcolor{white}{\textbf{Model}} & \textcolor{white}{\textbf{Single-1}} & \textcolor{white}{\textbf{Single-2}} & \textcolor{white}{\textbf{Single-3}} & \textcolor{white}{\textbf{MKey-1}} & \textcolor{white}{\textbf{MKey-2}} & \textcolor{white}{\textbf{MKey-3}} & \textcolor{white}{\textbf{MValue}} & \textcolor{white}{\textbf{MQuery}} & \textcolor{white}{\textbf{QA-1}} & \textcolor{white}{\textbf{QA-2}} & \textcolor{white}{\textbf{Avg}} \\
Qwen3.5-9B & 54.00 & 48.00 & 0.00 & 43.00 & 40.00 & 1.00 & 37.25 & 42.75 & 62.00 & 46.00 & 37.40 \\
\rowcolor{ResultStripe}
GLM-4.1V-9B & 61.00 & 26.00 & 0.00 & 30.00 & 19.00 & 0.00 & 19.50 & 33.25 & 50.00 & 48.00 & 28.68 \\
Glyph & 74.00 & 76.00 & 44.00 & 69.00 & 37.00 & 2.00 & 70.75 & 74.50 & 64.00 & 64.00 & 57.53 \\
\addlinespace[2pt]
\rowcolor{ResultFocus}
\textcolor{ResultTeal}{FocusVTC w/o GRPO} (Ours) & 50.00 & 39.00 & 0.00 & 29.00 & 37.00 & 0.00 & 24.25 & 24.50 & 65.00 & 54.00 & 32.28 \\
\rowcolor{ResultFocus}
\textcolor{ResultTeal}{FocusVTC w/o GRPO+tools} (Ours) & 43.00 & 29.00 & 1.00 & 24.00 & 37.00 & 1.00 & 23.75 & 23.00 & 22.00 & 26.00 & 22.98 \\
\rowcolor{ResultFocus}
\textcolor{ResultTeal}{FocusVTC-50} (Ours) & 82.99 & 82.73 & 46.03 & 73.50 & 66.15 & 37.59 & 54.36 & 71.10 & 77.27 & 68.58 & 66.03 \\
\rowcolor{ResultFocus}
\textcolor{ResultTeal}{FocusVTC-100} (Ours) & \underline{93.00} & \underline{96.00} & \underline{60.00} & \underline{87.00} & 75.00 & \textbf{49.00} & 63.50 & \underline{85.25} & 81.00 & \underline{73.00} & \underline{76.28} \\
\rowcolor{ResultFocus}
\textcolor{ResultTeal}{FocusVTC w/o tools} (Ours) & 51.00 & 45.00 & 0.00 & 38.00 & 42.00 & 1.00 & 30.00 & 30.00 & 64.00 & 65.00 & 36.60 \\
\rowcolor{ResultFocusStrong}
\textcolor{ResultTeal}{\textbf{FocusVTC}} (Ours) & \textbf{98.00} & \textbf{100.00} & \textbf{72.00} & \textbf{97.00} & \textbf{96.00} & \underline{46.00} & \textbf{96.25} & \textbf{98.50} & \textbf{89.00} & \textbf{81.00} & \textbf{87.38} \\
\addlinespace[2pt]
\rowcolor{ResultFocus}
\textcolor{ResultTeal}{FocusVTC w/o SFT} (Ours) & 85.53 & 85.26 & 51.59 & 81.69 & \underline{80.13} & 33.24 & \underline{79.53} & 82.70 & \underline{81.35} & 71.08 & 73.21 \\
Qwen3.5-9B+tools & 43.00 & 32.00 & 2.00 & 20.00 & 42.00 & 1.00 & 22.50 & 16.50 & 20.00 & 26.00 & 22.50 \\
\bottomrule
\end{tabular}}
\endgroup
\end{table}

\begin{table}[!htb]
\centering
\caption{RULER v2 task scores (\%) at 72 DPI. Avg is the unweighted arithmetic mean of the 12 displayed task scores. Best and second-best scores are bold and underlined, respectively. Qwen3.5-9B+tools uses tool access without additional training.}
\label{tab:ruler-v2-details}
\begingroup
\fontsize{8}{9.6}\selectfont
\setlength{\tabcolsep}{1.2pt}
\renewcommand{\arraystretch}{1.18}
\arrayrulecolor{ResultNavy}
\resizebox{\linewidth}{!}{%
\begin{tabular}{l*{13}{c}}
\rowcolor{ResultTableHeader}
\textcolor{white}{\textbf{Model}} & \multicolumn{4}{c}{\textcolor{white}{\textbf{MK-NIAH}}} & \multicolumn{4}{c}{\textcolor{white}{\textbf{MV-NIAH}}} & \multicolumn{4}{c}{\textcolor{white}{\textbf{QA}}} & \multicolumn{1}{c}{\textcolor{white}{\textbf{Overall}}} \\
\rowcolor{ResultHeader}
 & Basic & Easy & Medium & Hard & Basic & Easy & Medium & Hard & Basic & Easy & Medium & Hard & \textbf{Avg} \\
Qwen3.5-9B & 19.00 & 79.44 & 70.00 & 46.00 & 10.75 & 8.65 & 23.36 & 23.45 & 69.00 & 34.00 & 71.00 & 81.00 & 44.64 \\
\rowcolor{ResultStripe}
GLM-4.1V-9B & 8.00 & 51.12 & 29.00 & 25.00 & 3.50 & 9.67 & 10.64 & 18.24 & 53.00 & 32.00 & 53.58 & 53.50 & 28.94 \\
Glyph & 16.00 & 67.15 & 45.00 & 43.00 & 9.25 & 5.72 & 17.02 & 34.05 & 86.00 & \textbf{93.00} & 78.35 & 80.36 & 47.91 \\
\addlinespace[2pt]
\rowcolor{ResultFocus}
\textcolor{ResultTeal}{FocusVTC w/o GRPO} (Ours) & 18.00 & 59.96 & 51.00 & 39.00 & 7.25 & 7.05 & 16.79 & 27.24 & 74.00 & 83.00 & 68.50 & 71.17 & 43.58 \\
\rowcolor{ResultFocus}
\textcolor{ResultTeal}{FocusVTC w/o GRPO+tools} (Ours) & 20.00 & 71.32 & 45.00 & 33.00 & 10.75 & 7.03 & 13.04 & 29.78 & 72.00 & 36.00 & 63.27 & 71.68 & 39.41 \\
\rowcolor{ResultFocus}
\textcolor{ResultTeal}{FocusVTC-50} (Ours) & 56.82 & 90.05 & 75.48 & \underline{69.38} & 31.94 & 38.12 & \underline{51.07} & 54.98 & 90.88 & 82.16 & 84.11 & 77.57 & 66.88 \\
\rowcolor{ResultFocus}
\textcolor{ResultTeal}{FocusVTC-100} (Ours) & \underline{64.00} & \underline{95.61} & \underline{80.00} & \textbf{75.00} & 36.50 & \textbf{43.87} & \textbf{57.41} & \underline{60.11} & \underline{94.00} & 82.00 & \underline{87.00} & 78.75 & \underline{71.19} \\
\rowcolor{ResultFocus}
\textcolor{ResultTeal}{FocusVTC w/o tools} (Ours) & 22.00 & 78.68 & 48.00 & 40.00 & 10.00 & 9.25 & 17.09 & 27.49 & 84.00 & 56.00 & 70.77 & 74.58 & 44.82 \\
\rowcolor{ResultFocusStrong}
\textcolor{ResultTeal}{\textbf{FocusVTC}} (Ours) & \textbf{88.00} & \textbf{96.11} & \textbf{84.00} & \textbf{75.00} & \textbf{77.25} & \underline{39.72} & 35.59 & \textbf{61.79} & \textbf{95.00} & \underline{87.00} & \textbf{87.67} & \textbf{84.17} & \textbf{75.94} \\
\addlinespace[2pt]
\rowcolor{ResultFocus}
\textcolor{ResultTeal}{FocusVTC w/o SFT} (Ours) & 58.97 & 89.09 & 78.11 & 62.80 & \underline{49.27} & 26.65 & 30.44 & 45.66 & 84.06 & 64.70 & 80.65 & \underline{82.84} & 62.77 \\
Qwen3.5-9B+tools & 19.00 & 66.96 & 48.00 & 33.00 & 7.75 & 4.19 & 11.77 & 26.98 & 71.00 & 24.00 & 71.00 & 68.55 & 37.68 \\
\bottomrule
\end{tabular}}
\endgroup
\end{table}

\subsection{General Multimodal Capabilities}
\label{app:general-capabilities}

Table~\ref{tab:general-capabilities} lists the raw general multimodal scores
from Figure~\ref{fig:introduction}(d), including Glyph and GLM-4.1V-9B,
followed by the MMLongBench comparison. \model improves over Qwen3.5-9B
on all six general benchmarks and on MMLongBench image and text needle retrieval,
while the remaining MMLongBench results are comparable overall.

\begin{table}[H]
\caption{General multimodal scores from Figure~\ref{fig:introduction}(d)
(top) and MMLongBench results (bottom). DocVQA and InfoVQA use ANLS;
other entries retain their benchmark scales.}
\label{tab:general-capabilities}
\centering
\begingroup
\fontsize{8}{9.6}\selectfont
\setlength{\tabcolsep}{3pt}
\renewcommand{\arraystretch}{1.18}
\arrayrulecolor{ResultNavy}
\resizebox{\linewidth}{!}{%
\begin{tabular}{l*{6}{c}}
\rowcolor{ResultTableHeader}
\textcolor{white}{\textbf{Model}} & \textcolor{white}{\textbf{OCRBench}} & \textcolor{white}{\textbf{DocVQA}} & \textcolor{white}{\textbf{MMMU}} & \textcolor{white}{\textbf{MME}} & \textcolor{white}{\textbf{ChartQA}} & \textcolor{white}{\textbf{InfoVQA}} \\
Qwen3.5-9B~\citep{qwen2026qwen35} & 851 & 92.38 & 65.12 & 2424.02 & 85.96 & 74.76 \\
\rowcolor{ResultStripe}
Glyph~\citep{cheng2025glyph} & 799 & 91.75 & 57.67 & 2253.13 & 72.76 & 71.89 \\
GLM-4.1V-9B~\citep{vteam2025glm45vglm41vthinkingversatilemultimodal} & 820 & 91.34 & 64.33 & 2392.53 & 70.76 & 70.52 \\
\rowcolor{ResultFocusStrong}
\textcolor{ResultTeal}{\textbf{FocusVTC}} (Ours) & 860 & 92.43 & 66.73 & 2457.62 & 86.28 & 76.20 \\
\specialrule{0.45pt}{5pt}{5pt}
\rowcolor{ResultTableHeader}
\textcolor{white}{\textbf{Model}} & \textcolor{white}{\textbf{VRAG}} & \textcolor{white}{\textbf{VH}} & \textcolor{white}{\textbf{NIAH image}} & \textcolor{white}{\textbf{NIAH text}} & \textcolor{white}{\textbf{Summ}} & \textcolor{white}{\textbf{DocQA}} \\
Qwen3.5-9B~\citep{qwen2026qwen35} & 61.12 & 56.10 & 45.04 & 71.82 & 29.19 & 68.96 \\
\rowcolor{ResultFocusStrong}
\textcolor{ResultTeal}{\textbf{FocusVTC}} (Ours) & 61.00 & 57.05 & 49.82 & 76.11 & 28.40 & 67.27 \\
\bottomrule
\end{tabular}%
}
\endgroup
\end{table}

\section{Limitations}
\label{sec:limitations}

Despite promising improvements in the compression--performance balance,
our current study is limited in model and training-data scale, which defines
several directions for future research. We develop \model with a 9B-parameter
multimodal backbone and 29.4K verified REL-CoT examples. While this setting
demonstrates the effectiveness of adaptive-resolution reading, the benefits
of scaling model capacity and localization supervision remain to be
established. Moreover, long-context compression must accommodate diverse
languages, document layouts, and reasoning demands; the current training
corpus represents only part of this broader space. Future work will therefore
focus on extending the framework to larger backbones and expanding REL-CoT
with more diverse documents and reasoning trajectories. We also plan to
study adaptive enhancement at larger context scales, jointly evaluating
answer quality and complete interaction cost. These
extensions would help establish the scalability of selective visual reading
and broaden its applicability while preserving general multimodal
capabilities.

\clearpage
\endgroup
\section{Case Studies}
\label{app:case-studies}

\subsection{Reasoning with Page and Region Annotations}
\label{app:rel-case-study}

\begin{figure}[!ht]
  \centering
  \includegraphics[width=\linewidth]{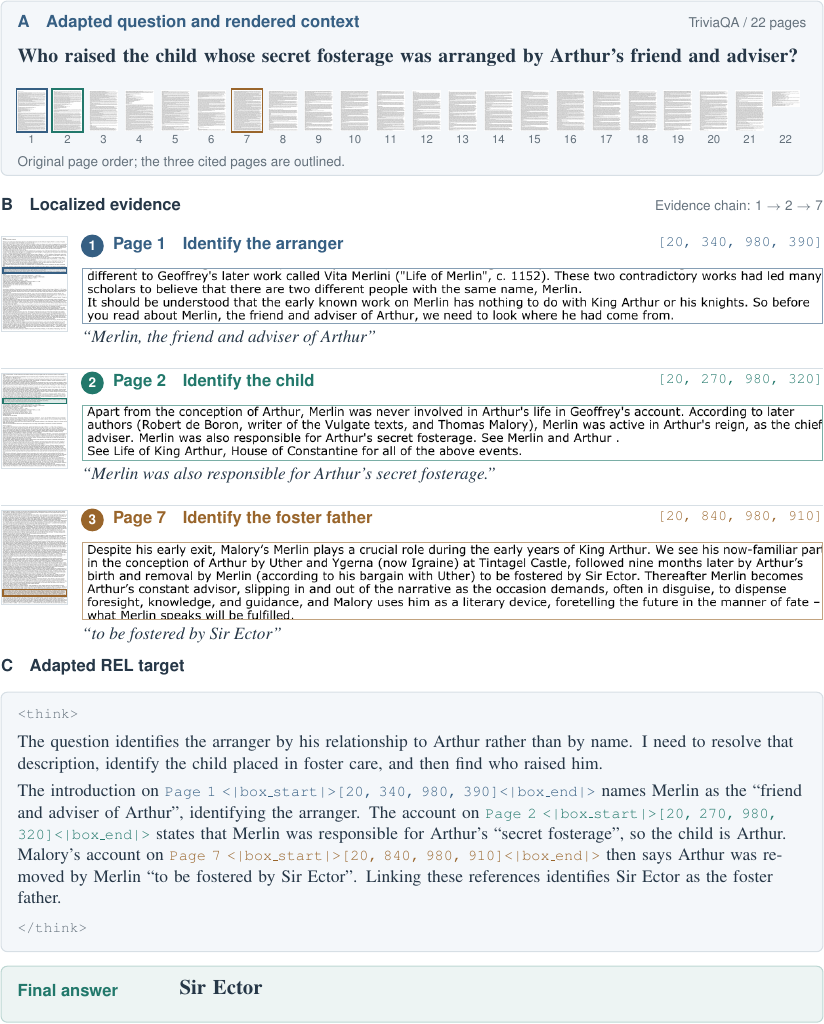}

  \caption{\textbf{A factual evidence chain in REL-CoT.}
  The question and target are adapted; source pages and boxes are unchanged.
  (A)~A factual question over 22 pages (72-DPI thumbnails).
  (B)~144-DPI excerpts linking Merlin, Arthur, and Sir Ector; box coordinates
  are normalized to $[0,1000]$.
  (C)~Initial question analysis followed by sentence-embedded page--box
  citations and a short, directly verifiable answer.}
  \label{fig:rel-case-study}
\end{figure}

Figure~\ref{fig:rel-case-study} follows an annotated reasoning trace through
Merlin and Arthur to the answer, Sir Ector.

\clearpage

\subsection{Adaptive Reading Across Pages}
\label{app:case-study}

\begin{figure}[!ht]
  \centering
  \includegraphics[width=\linewidth]{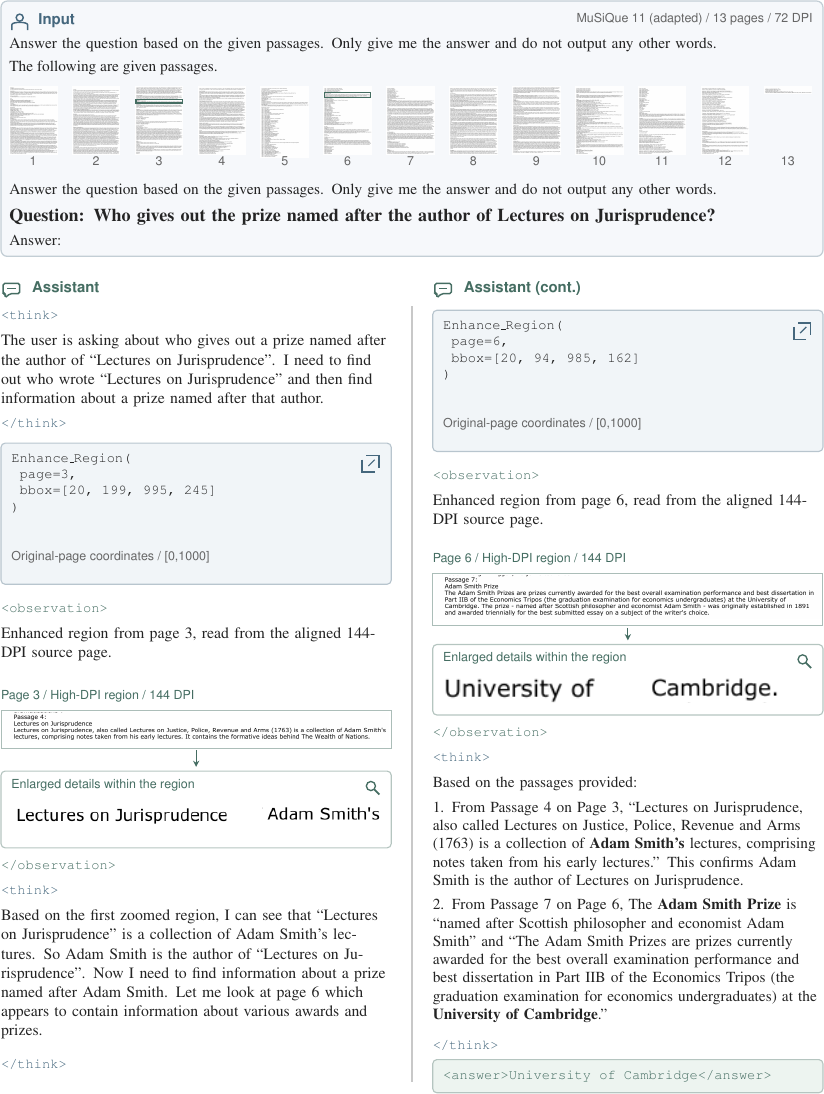}
  \caption{\textbf{An adapted MuSiQue illustration of selective enhancement.}
  The question, source pages, reasoning excerpts, and answer are retained
  from the recorded Verdana example; tool-call boxes and observations are
  adapted for illustration. \texttt{Enhance\_Region} reads local passages on
  pages 3 and 6 from aligned 144-DPI sources, using original-page coordinates
  in $[0,1000]$. Insets enlarge text within these enhanced regions.}
  \label{fig:case-study}
\end{figure}

\end{document}